%% file: causal_sa.tex
\documentclass[sigconf]{acmart}
\AtBeginDocument{%
	\providecommand\BibTeX{{%
			\normalfont B\kern-0.5em{\scshape i\kern-0.25em b}\kern-0.8em\TeX}}}

\copyrightyear{2021}
\acmYear{2021}
\setcopyright{rightsretained}
\acmConference[ACM CHIL '21]{ACM Conference on Health, Inference,
and Learning}{April 8--10, 2021}{Virtual Event, USA}
\acmBooktitle{ACM Conference on Health, Inference, and Learning (ACM
CHIL '21), April 8--10, 2021, Virtual Event,
USA}\acmDOI{10.1145/3450439.3451875}
\acmISBN{978-1-4503-8359-2/21/04}

\usepackage{hyperref}
\usepackage{url}

\usepackage[utf8]{inputenc} 
\usepackage[T1]{fontenc}    
\usepackage{hyperref}       
\usepackage{url}            
\usepackage{booktabs}       
\usepackage{amsfonts}       
\usepackage{nicefrac}       
\usepackage{microtype}      

\usepackage{amsmath}
\usepackage[export]{adjustbox}
\usepackage{arydshln} 
\usepackage{wrapfig}
\usepackage{color}
\usepackage{multirow}
\usepackage{subcaption}
\usepackage{soul}
\usepackage{enumitem}

\newtheorem{corollary}{Corollary}

\newtheorem{definition}{Definition}

\newcommand{\CD}{\mathcal{D}}

\newcommand{\epsilonv}{\boldsymbol{\epsilon}}
\newcommand{\bigCI}{\mathrel{\text{\scalebox{1.07}{$\perp\mkern-10mu\perp$}}}}

\LetLtxMacro{\originaleqref}{\eqref}
\renewcommand{\eqref}[1]{(\ref{#1})}

\begin{document}
	
	\title{Enabling Counterfactual Survival Analysis \\ with Balanced Representations}
	

	\author{Paidamoyo Chapfuwa}
	\affiliation{
		\institution{Duke University}
			\country{USA}
	}
	\email{paidamoyo.chapfuwa@duke.edu}

	\author{Serge Assaad}
	\affiliation{
		\institution{Duke University}
			\country{USA}
	}
	\email{serge.assaad@duke.edu}

	\author{Shuxi Zeng}
	\affiliation{
		\institution{Duke University}
		\country{USA}
	}
	\email{zengshx777@gmail.com}
	
		\author{Michael J. Pencina}
	\affiliation{
		\institution{Duke University}
			\country{USA}
	}
	\email{michal.pencina@duke.edu}

	\author{Lawrence Carin}
	\affiliation{
		\institution{Duke University}
			\country{USA}
	}
	\email{lcarin@duke.edu}
	
	\author{Ricardo Henao}
	\affiliation{\institution{Duke University}
		\country{USA}
	}
	\email{ricardo.henao@duke.edu}
	
	\renewcommand{\shortauthors}{Chapfuwa, et al.}
	
	\begin{abstract}
		A clear and well-documented \LaTeX\ document is presented as an
		article formatted for publication by ACM in a conference proceedings
		or journal publication. Based on the ``acmart'' document class, this
		article presents and explains many of the common variations, as well
		as many of the formatting elements an author may use in the
		preparation of the documentation of their work.
	\end{abstract}

	
	
	\begin{abstract}
		Balanced representation learning methods have been applied successfully to counterfactual inference from observational data.
		However, approaches that account for survival outcomes are relatively limited.
		Survival data are frequently encountered across diverse medical applications, \textit{i.e.}, drug development, risk profiling, and clinical trials, and such data are also relevant in fields like manufacturing (\textit{e.g.}, for equipment monitoring).
		When the outcome of interest is a time-to-event, special precautions for handling censored events need to be taken, as ignoring censored outcomes may lead to biased estimates. 
		We propose a theoretically grounded unified framework for counterfactual inference applicable to survival outcomes.
		Further, we formulate a nonparametric hazard ratio metric for evaluating average and individualized treatment effects.
		Experimental results on real-world and semi-synthetic datasets, the latter of which we introduce, demonstrate that the proposed approach significantly outperforms competitive alternatives in both survival-outcome prediction and treatment-effect estimation.
	\end{abstract}
	
	\maketitle

	\section{INTRODUCTION}
	
	Survival analysis or time-to-event studies focus on modeling the time of a future event, such as death or failure, and investigate its relationship with covariates or predictors of interest.
	Specifically, we may be interested in the \textit{causal effect} of a given intervention or treatment on survival time.
	A typical question may be: will a given therapy increase the chances of survival of an individual or population?
	Such causal inquiries on survival outcomes are common in the fields of epidemiology and medicine \citep{robins1986new,hammer1996trial, yusuf2016cholesterol}.
	As an important current example, the COVID-19 pandemic is creating a demand for methodological development to address such questions, specifically, when evaluating the effectiveness of a potential vaccine or therapeutic outside randomized controlled trial settings.
	
	Traditional causal survival analysis is typically carried out in the context of a randomized controlled trial (RCT), where the treatment assignment is controlled by researchers.
	Though they are the gold standard for causal inference, RCTs are usually long-term engagements, expensive and limited in sample size.
	Alternatively, the availability of \textit{observational} data with comprehensive information about patients, such as electronic health records (EHRs), constitutes a more accessible but also more challenging source for estimating causal effects \citep{hayrinen2008definitionEHR,jha2009useEHR}. Such observational data may be used to augment and verify an RCT, after a particular treatment is approved and in use \citep{gombar2019time, frankovich2011evidence, longhurst2014green}. 
	Moreover, the wealth of information from observational data also allows for the estimation of the individualized treatment effect (ITE), namely, the causal effect of an intervention at the individual level.
	In this work, we develop a novel framework for {\em counterfactual time-to-event prediction} to estimate the ITE for survival or time-to-event outcomes from observational data.
	
	Estimating the causal effect for survival outcomes in observational data manifests two principal challenges.
	First, the treatment assignment mechanism is not known \textit{a priori}.
	Therefore, there may be variables, known as \textit{confounders}, affecting both the treatment and survival time, which lead to selection bias \citep{bareinboim2012controlling}, \textit{i.e.}, that the distributions across treatment groups are not the same. In this work, we focus on selection biases due to confounding, but other sources may also be considered.
	For instance, patients who are severely ill are likely to receive more aggressive therapy, however, their health status may \textit{also} inevitably influence survival.
	Traditional survival analysis neglects such bias, leading to incorrect causal estimation.
	Second, the exact time-to-event is not always observed, \textit{i.e.}, sometimes we only know that an event has \textit{not} occurred up to a certain point in time.
	This is known as the \textit{censoring} problem.
	Moreover, censoring might be informative depending on the characteristics of the individuals and their treatment assignments, thus proper adjustment is required for accurate causal estimation \citep{cole2004adjusted, diaz2019statistical}.
	
	Traditional causal survival-analysis approaches typically model the effect of the treatment or covariates (not time or survival) in a parametric manner.
	Two commonly used models are the Cox proportional hazards (CoxPH) model \citep{cox1972regression} and the accelerated failure time (AFT) model \citep{wei1992accelerated}, which presume a linear relationship between the covariates and survival probability. 
	Further, proper weighting for each individual has been employed to account for confounding bias from these models \citep{austin2007propensity,austin2014use,hernan2005structuralaft}.
	For instance, probability weighting schemes that account for both selection bias and covariate dependent censoring have been considered for adjusted survival curves \citep{cole2004adjusted, diaz2019statistical}.
	Moreover, such probability weighting schemes have been applied to causal survival-analysis under time-varying treatment and confounding \citep{robins1986new, hernan2000marginal}.
	See \citet{van2003unified, tsiatis2007semiparametric, van2011targeted, hernan2020causal} for an overview.
	Such linear specification makes these models interpretable but compromises their flexibility, and makes it difficult to adapt them for high-dimensional data or to capture complex interactions among covariates. 
	Importantly, these methods lack a counterfactual prediction mechanism, which is key for ITE estimation (see Section 2).
	
	Fortunately, recent advances in machine learning, such as representation learning or generative modeling, have enabled causal inference methods to handle high-dimensional data and to characterize complex interactions effectively.
	For instance, there has been recent interest in tree-based \citep{chipman2010bart,wager2018estimation} and neural-network-based \citep{shalit2017estimating,zhang2020learning} approaches.
	For pre-specified time-horizons, the nonparametric Random Survival Forest (RSF) \citep{ishwaran2008random} and Bayesian Additive regression trees (BART) \citep{chipman2010bart} have been extended to causal survival analysis.
	RSF has been applied to causal survival forests with weighted bootstrap inference  \citep{shen2018estimating, cui2020estimating}  while a BART  is extended to account for survival outcomes in Surv-BART \citep{sparapani2016nonparametric},  and AFT-BART \citep{henderson2020individualized}. See \cite{hu2020estimating} for an extensive investigation of  the causal survival  tree-based methods.
	
	Alternatively, when estimating the ITE, neural-network-based methods propose to regularize the transformed covariates or representations for an individual to have balanced distributions across treatment groups, thus accounting for the confounding bias and improving ITE prediction.
	However, most approaches employing {\em representation learning} techniques for counterfactual inference deal with continuous or binary outcomes, instead of time-to-event outcomes with censoring (\emph{informative or non-informative}). Moreover, while recent neural-network-based survival analysis methods \citep{nagpal2021deep, ranganath2016deep, chapfuwa2018adversarial, avati2018countdown, miscouridou2018deep,lee2019temporal, lee2018deephit, xiu2020variational} have improved survival predictions when censoring is non-informative, they lack mechanisms for accounting for informative censoring or confounding biases.
	Hence, a principled generalization to the context of {\em counterfactual survival analysis} is needed.
	
	In this work we leverage balanced (latent) representation learning to estimate ITEs via counterfactual prediction of survival outcomes in observational studies.
	We develop a framework to predict event times from a low-dimensional transformation of the original covariate space.
	To address the specific challenges associated with counterfactual survival analysis, we make the following contributions:
	%
	\begin{itemize}[topsep=9pt,itemsep=0pt,parsep=0pt,partopsep=0pt,leftmargin=10pt]
		\item We develop an optimization objective incorporating adjustments for informative censoring, as well as a balanced regularization term bounding the generalization error for ITE prediction.
		For the latter, we repurpose a recently proposed bound \citep{shalit2017estimating} for our time-to-event scenario.
		\item We propose a generative model for event times to relax restrictive survival linear and parametric assumptions, thus allowing for more flexible modeling.
		Our approach can also provide nonparametric uncertainty quantification for ITE predictions.
		\item We provide survival-specific evaluation metrics, including a new \textit{nonparametric hazard ratio} estimator, and discuss how to perform model selection for survival outcomes.
		The proposed model demonstrates superior performance relative to the commonly used baselines in real-world and semi-synthetic datasets.
		\item We introduce a survival-specific semi-synthetic dataset and demonstrate an approach for leveraging prior randomized experiments in longitudinal studies for model validation.
	\end{itemize}
	
	\section{Problem Formulation}
	
	We first introduce the basic setup for performing causal survival analysis in observational studies.
	Suppose we have $N$ units, with $N_{1}$ units being \textit{treated} and $N_{0}$ in the \textit{control} group ($N=N_1+N_0$).
	For each unit (individual), we have covariates $X$, which can be heterogeneous, \textit{e.g.}, a mixture of categorical and continuous covariates which, in the context of medicine, may include labs, vitals, procedure codes, \textit{etc}.
	We also have a \emph{treatment} indicator $A$, where $A=0$ for the controls and $A=1$ for the treated, as well as the outcome (event) of interest $T$.
	Under the potential-outcomes framework \citep{Rubin_potential_outcomes}, let $T_{0}$ and $T_{1}$ be the potential event times for a given subject under control and treatment, respectively.
	In practice we only observe one realization of the potential outcomes, \textit{i.e.}, the \textit{factual} outcome $T =T_{A}$, while the \textit{counterfactual} outcome $T_{1-A}$ is unobserved.
	
	In survival analysis, the problem becomes more difficult because we do \textit{not} always observe the exact event time for each individual, but rather the time up to which we are certain that the event has not occurred; specifically, we have a (right) censoring problem, most likely due to the loss of follow-up.
	We denote the censoring time as $C$ and censoring indicator as $\delta \in \{0,1\}$.
	The actual \textit{observed time} is $Y =\min (T_A ,C )$, \textit{i.e.}, the outcome is observed (non-censored) if $T_A< C$ and $\delta =1$.
	
\begin{figure*}[t]
		\begin{subfigure}{.48\textwidth}
			\centering
			\includegraphics[width=.8\linewidth]{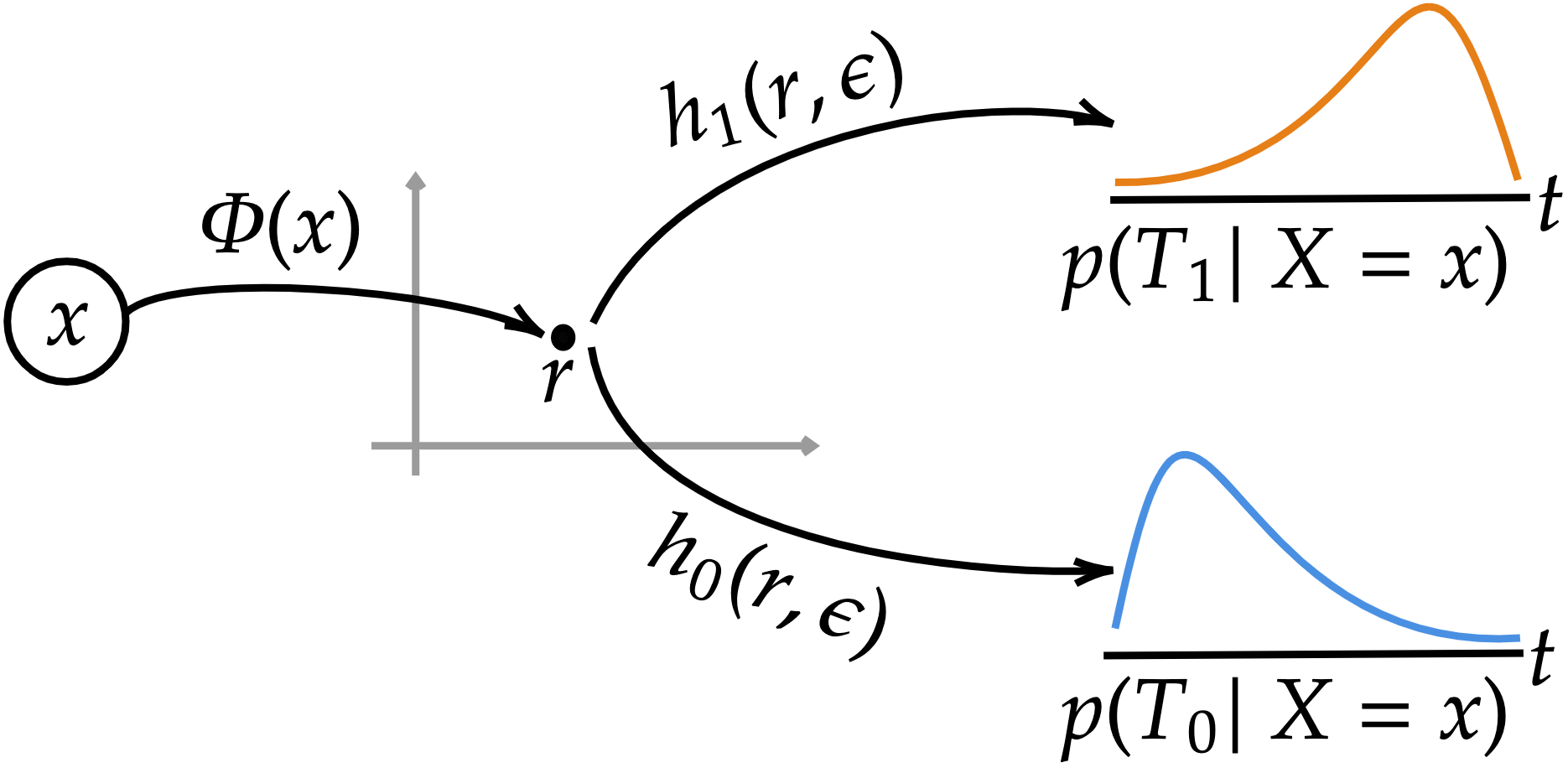}  
			\caption{Counterfactual inference}
			\label{fig:model}
		\end{subfigure}
		\begin{subfigure}{.25\textwidth}
			\centering
			\includegraphics[width=.65\linewidth]{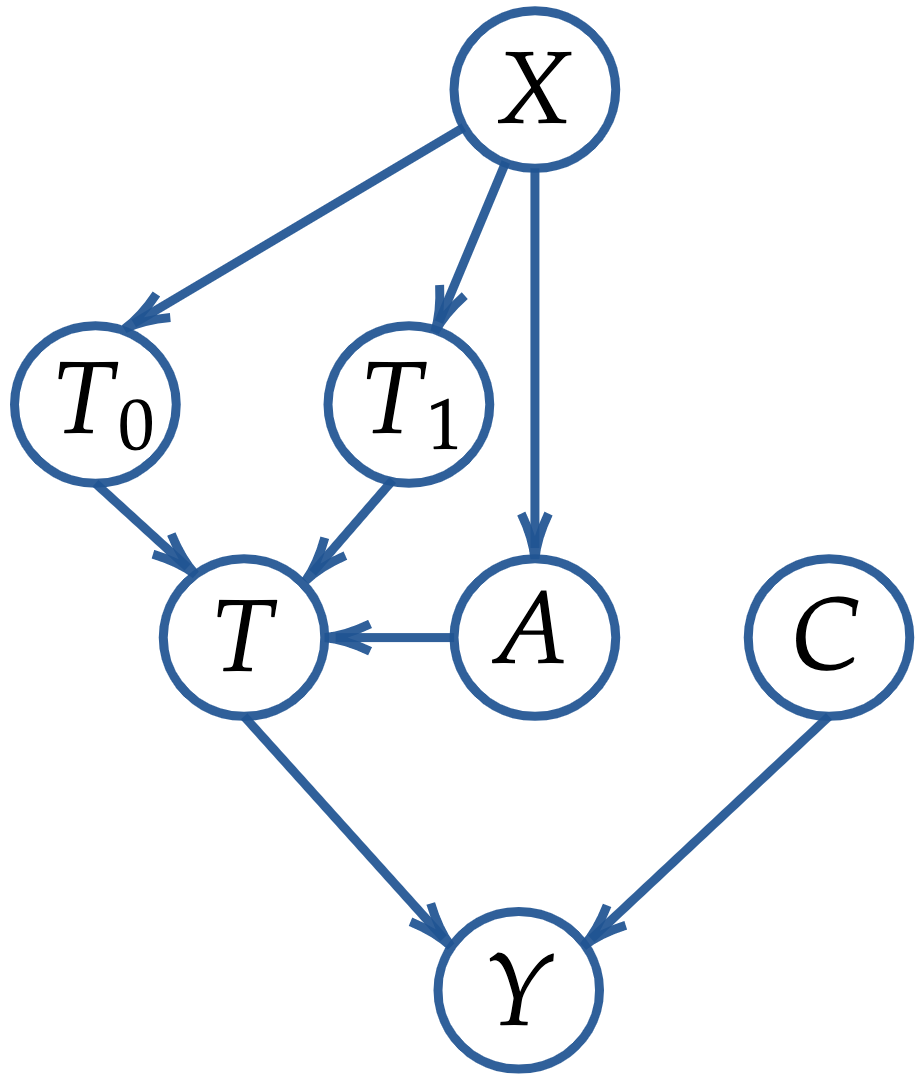}
			\caption{Non-informative}
			\label{fig:non_informative}
		\end{subfigure}
		\begin{subfigure}{.24\textwidth}
			\centering
			\includegraphics[width=.7\linewidth]{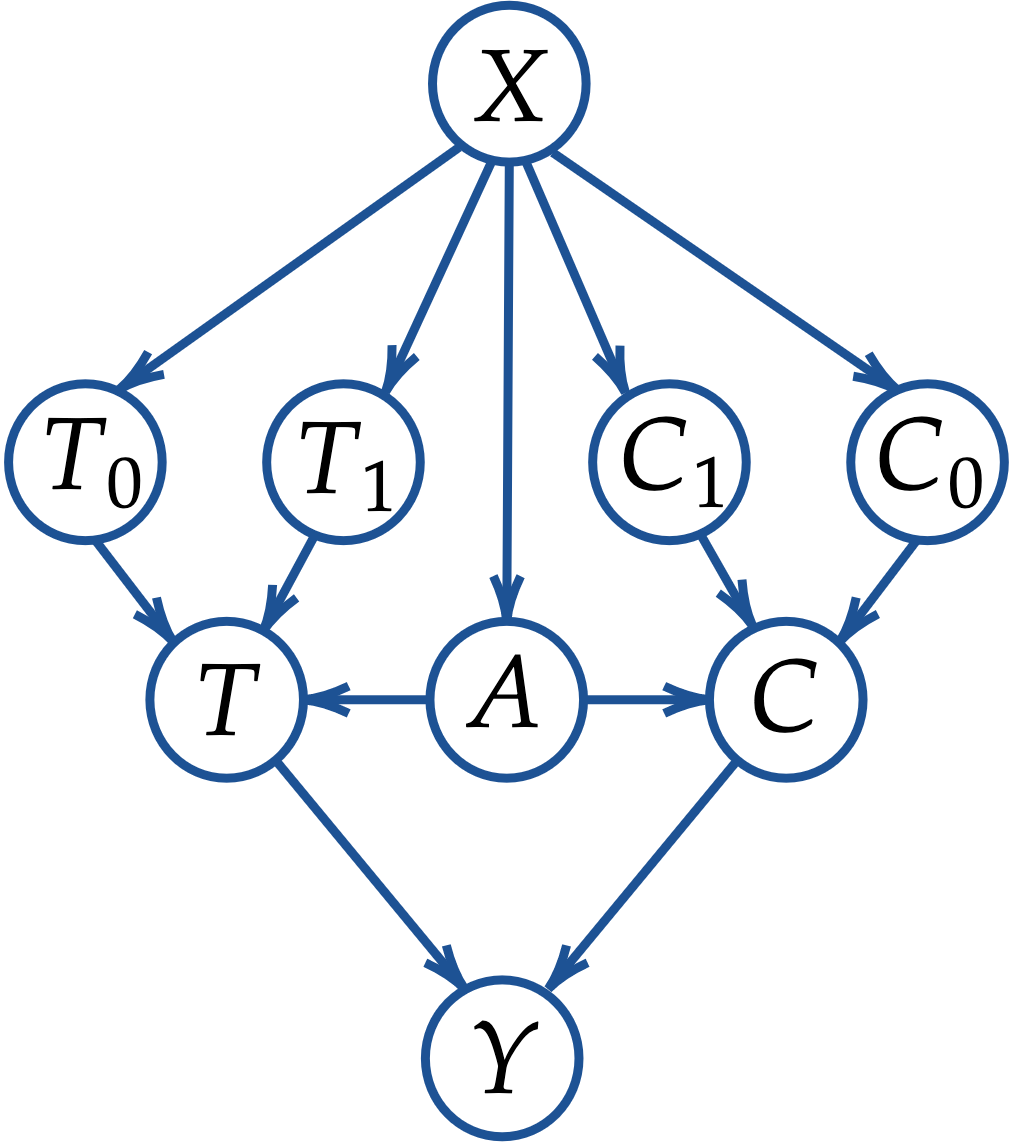}
			\caption{Informative}
			\label{fig:informative}
		\end{subfigure}
		\caption{(a) Illustration of the proposed counterfactual survival analysis (CSA). Covariates $X=x$ are mapped into latent representation $r$ via deterministic mapping $r=\Phi(x)$. The potential outcomes are sampled from $t_a \sim p(T_{A} |X= x)$ for $A=a$ via stochastic mapping $h_{A}({r},\tilde{\epsilon})$, where stochasticity is induced with a planar-flow-based transformation, $\tilde{\epsilon}$, of a simple distribution $p(\epsilon)$, \textit{i.e.}, uniform or Gaussian. (b) and (c) show the proposed causal graphs for non-informative and informative censoring, respectively.}
		\label{fig:causal_graphs}
	\end{figure*}

	In this work, we are interested in the expected difference between the $T_1$ and $T_0$ conditioned on $X$ for a given unit (individual), which is commonly known as the \textit{individualized treatment effect} (ITE).
	Specifically, we wish to perform inference on the conditional distributions of $T_1$ and $T_0$, \textit{i.e.}, $p(T_1|X)$ and $p(T_0|X)$, respectively, as shown in Figure~\ref{fig:model}. In practice, we observe $N$ realizations of $(Y, \delta, X, A)$ for observed time, censoring indicator, covariates and treatment indicator, respectively; hence, from an observational study the dataset takes the form $\CD = \{(y_i, \delta_i, {x}_i, a_i)\}_{i=1}^N$.
	Below, we discuss several common choices of estimands in survival analysis.
	
	\subsection{Estimands of Interest}\label{sec:estimands}
	%
	We begin by considering survival analysis in the {\em absence} of an intervening treatment choice, $A$.
	Let $F(t|x) \triangleq P(T\leq t |X=x)$ be the cumulative distribution function of the event (failure) time, $t$, given a realization of the covariates, $x$.
	Survival analysis is primarily concerned with characterization of the \textit{survival function} conditioned on covariates $S(t|x) \triangleq 1-F(t|x)$, and the \emph{hazard function} or \emph{risk score}, $\lambda(t|x)$, defined below.
	$S(t|x)$ is a monotonically decreasing function indicating the probability of survival up to time $t$.
	The hazard function measures the instantaneous probability of the event occurring between $\{t , t +  \Delta t\} $ given $T > t$ and $\Delta t\to 0$.
	From standard definitions \citep{kleinbaum2010survival}, the relationship between cumulative and hazard function is formulated as 
	\begin{align}\label{eq:hazard}
	\begin{aligned}
	\lambda(t|x)& = \lim\limits_{dt \rightarrow 0}  \frac{P(t <T < t + dt|X=x)}{P(T > t|X=x) dt} \\
	& = -\frac{d \log S(t|x)}{dt}=\frac{f(t|x)}{S(t|x)}.
	\end{aligned}
	\end{align}
	From \eqref{eq:hazard} we see that $f(t|x)\triangleq P(T=t|X=x) = \lambda(t|x)S(t|x)$, is the conditional \emph{event time density function} \citep{kleinbaum2010survival}.
	
	Given the binary treatment $A$, we are interested in its impact on the survival time. For ITE estimation, we are also interested in the difference between the two potential outcomes $T_1,T_0$.
	Let $S_{A}(t|x)$ and $\lambda_{A}(t|x)$ denote the survival and hazard functions for the potential outcomes $T_A$, \textit{i.e.}, $T_1$ and $T_0$.
	Several common estimands of interest include \citep{zhao2012utilizing, trinquart2016comparison}:
	
	\begin{itemize}[topsep=0pt,itemsep=0pt,parsep=0pt,partopsep=0pt,leftmargin=10pt]
		\item {\em Difference in expected lifetime}: \newline
		$\textup{ITE}(x)=\int_{0}^{t_{\max}}\{S_{1}( t|x)-S_{0}(t|x)\}\textup{d}t=\mathbb{E}\{T_1 -T_0|X =x\}$.
		\item {\em Difference in survival function}: $\textup{ITE}(t, x)=S_{1}(t |x)-S_{0}(t |x)$.
		\item {\em Hazard ratio}: $\textup{ITE}(t,x)=\lambda_{1}(t |x)/\lambda_{0}(t|x)$.
	\end{itemize}

	The inference difficulties associated with the above estimands from observational data are two-fold.
	First, there are confounders affecting both the treatment assignment and outcomes, which stem from selection bias, \textit{i.e.}, the treatment and control covariate distributions are not necessarily the same.
	Also, we do not have direct knowledge of the conditional treatment assignment mechanism, \textit{i.e.}, $P(A =a|X =x)$, also known as the \textit{propensity score}.
	Let $\bigCI$ denote statistical independence.
	For estimands to be identifiable from observational data, we make two assumptions: 
	($i$) $\{T_1, T_0 \} \bigCI A | X $, \textit{i.e.}, no unobserved confounders or \textit{ignorability}, and 
	($ii$) \emph{overlap} in the covariate support $0< P(A =1|X=x)<1 $ \emph{almost surely if} $p(X=x) > 0$.
	Second, the censoring mechanism is also unknown and may lead to bias without proper adjustment.
	We consider two censoring mechanisms in our work, 
	($i$) conditionally independent or \textit{informative censoring}: $T  \bigCI C | X , A$, and 
	($ii$) random or \textit{non-informative censoring}: $T \bigCI C$.
	Note that for informative censoring, we also have to consider potential censoring times $C_1$ and $C_0$ and their conditionals $p(C_1|X)$ and $p(C_0|X)$, respectively.
	Figure \ref{fig:causal_graphs} shows causal graphs illustrating these modeling assumptions.
	
	\vspace{-1mm}
	\section{Modeling}
	To overcome the above challenges and adjust for observational biases, we propose a unified framework for {\em counterfactual survival analysis} (CSA).
	Specifically, we repurpose the counterfactual bound in \citet{shalit2017estimating} for our time-to-event scenario and introduce a nonparametric approach for stochastic survival outcome predictions.
	Below we formulate a theoretically grounded and unified approach for estimating 
	$(i)$ the encoder function ${r}=\Phi(x)$, which  deterministically maps covariates $x$ to their corresponding latent representation ${r}\in\mathbb{R}^d$, and 
	$(ii)$ two stochastic time-to-event generative functions, $h_A(\cdot)$, to implicitly draw samples from both potential outcome conditionals $t_a \sim p_{h, \Phi} (T_A |X=x)$, for $A=\{1,0\}$, and where $t_a$ indicates the sample from $p_{h, \Phi} (T_A |X=x)$ is for $A=a$.
	Further, we formulate a general extension that accounts for informative censoring by introducing two stochastic censoring generative functions, $\nu_A (\cdot)$, to draw samples for potential censoring times $c_a \sim p_{\nu,\Phi } (C_A |X=x)$.
	The model-specifying functions, $\{h_A(\cdot), \nu_A(\cdot), \Phi(\cdot)\}$, are parameterized via neural networks.
	See the Supplementary Material (SM) for details.
	Figure \ref{fig:model} summarizes our modeling approach.
	
	\subsection{Accounting for selection bias}
	%
	We wish to estimate the potential outcomes, \textit{i.e.}, event times, which are sampled by distributions parameterized by functions $\{h_A(\cdot),\Phi(\cdot)\}$, \emph{i.e.}, 
	\begin{align}
	\label{eq:identifiable}
	t & \sim  p_{h, \Phi} (T |X=x, A=a) 
	\\
	t_a & \sim p_{h, \Phi} (T_a |X=x)
	\label{eq:ignorability}
	\end{align}
	We obtain \eqref{eq:ignorability} from \eqref{eq:identifiable} via the \emph{strong ignorability} assumption, \textit{i.e.}, $ \{T_0,T_1\} \bigCI A | X$ (consistent with the causal graphs in Figure~\ref{fig:non_informative} and \ref{fig:informative}) and $ 0 < P(A=a | X=x) < 1$, and the \emph{consistency} assumption, \textit{i.e.}, $T = T_A|A=a$.
	A similar argument can be made for informative censoring based on Figure~\ref{fig:informative}, so we can also write $c_a \sim p_{\nu,\Phi } (C_A |X=x)$.
	Given~\eqref{eq:ignorability}, model functions $\{h_A(\cdot),\Phi(\cdot)\}$ and $\nu_A(\cdot)$ for informative censoring can be learned by leveraging standard statistical optimization approaches, that minimize a loss hypothesis $\mathcal{L}$ given samples from the empirical distribution $(y,\delta,x,a) \sim p(Y,\delta,X,A)$, \textit{i.e.}, from dataset $\CD$.
	Specifically, we write $\mathcal{L}$ as
	\begin{align}\label{eq:loss}
	\mathcal{L} & = \mathbb{E}_{(y,\delta,x,a) \sim p(Y,\delta,X,A)} \left[
	\ell_{h, \Phi} ( t_a, y, \delta) \right] \,,
	\end{align}
	%
	where $\ell_{h, \Phi} ( t_a, y, \delta)$ is a loss function that measures the agreement of $t_a \sim p_{h, \Phi} (T_A |X=x)$ (and $c_a \sim p_{\nu,\Phi } (C_A |X=x)$ for informative censoring) with ground truth $\{y,\delta\}$, the observed time and censoring indicator, respectively.
	
	For some parametric formulations of event time distribution $ p_{h, \Phi} (T_A |X=x)$, \textit{e.g.}, exponential, Weibull, log-Normal, \textit{etc}., and provided the censoring mechanism is non-informative, $-\ell_{h, \Phi} ( t_a, y, \delta)$ is the closed form  log likelihood.
	Specifically, $ -\ell_{h, \Phi} (t_a, y, \delta) \triangleq \log p_{h,\Phi}(T_a|X=x) = \delta \cdot \log f_{h,\Phi}(t_a|x) + (1-\delta) \cdot  \log S_{h,\Phi}(t_a|x) $, which implies that the conditional event time density and survival functions can be calculated in closed form from transformations $\{h_A(\cdot),\Phi(\cdot)\}$ of $x$.
	See the SM for parametric examples of \eqref{eq:loss} accounting for informative censoring.
	
	We further define the expected loss for a given realization of covariates $x$ and treatment assignment $a$ over observed times $y$ (censored and non-censored), and the censoring indicator $\delta$ as
	%
	$\zeta_{h,\Phi}(x,a) \triangleq \mathbb{E}_{(y,\delta,x) \sim p(Y,\delta|X)} \ell_{h, \Phi} (t_a, y, \delta)$ 
	%
	as in \citet{shalit2017estimating}.
	%
	For a given subject with covariates $x$ and treatment assignment $a$, we wish to minimize both the factual and counterfactual losses, $\mathcal{L}_{\rm F}$ and $\mathcal{L}_{\rm CF}$, respectively, by decomposing $\mathcal{L} = \mathcal{L}_{\rm F} + \mathcal{L}_{\rm CF}$ as follows
	\begin{align}\label{eq:f_and_cf}
	\begin{aligned}
	\mathcal{L}_{\rm F} &=  \mathbb{E}_{(x, a) \sim p( A, X)} 	\zeta_{h,\Phi}(x,a)\,,  \\
	\mathcal{L}_{\rm CF} &= \mathbb{E}_{(x, a) \sim p( 1-A, X)}	\zeta_{h,\Phi}(x,a)\,.
	\end{aligned}
	\end{align}
	Let $u \triangleq P(A=1)$ denote the marginal probability of treatment assignment.
	We can readily decompose the losses in \eqref{eq:f_and_cf} according to treatment assignments.
	The decomposed factual $\mathcal{L}_{\rm F} = u\cdot \mathcal{L}_{\rm F} ^{A=1} + (1-u)\cdot\mathcal{L}_{\rm F}^{A=0}$, and similarly, the decomposed counterfactual $\mathcal{L}_{\rm CF} = (1-u)\cdot \mathcal{L}_{\rm CF}^{A=1} + u\cdot\mathcal{L}_{ \rm CF}^{A=0}$.
	In practice, only \emph{factual} outcomes are observed, hence, for a non-randomized non-controlled experiment, we cannot obtain an unbiased estimate of $\mathcal{L}_{\rm CF}$ from data due to selection bias (or confounding).
	Therefore, we bound $\mathcal{L}_{\rm CF}$ and $\mathcal{L}$ below following \citet{shalit2017estimating}.
	\begin{corollary}
		Assume \ $\Phi(\cdot)$ is an invertible map, and \ $ \alpha^{-1}\zeta_{h,\Phi}(x,a)  \in G$, where $G$ is a family of functions, $p_{\Phi}^{A=a} \triangleq p_{\Phi} ( R | A=a)$ is the latent distribution for group $A=a$, and $\alpha >0$ is a constant. Then, we have: 
		%
		\begin{align} 
		\mathcal{L}_{\rm CF} & \le (1-u)\cdot \mathcal{L}_{\rm F}^{A=1} + u \cdot \mathcal{L}_{\rm F}^{A=0} +  \alpha\cdot  {\rm IPM}_{G}(p_\Phi^{A=1},p_\Phi^{A=0}) \notag\\
		\label{eq:upper_bound}
		\mathcal{L} & \le \mathcal{L}_{\rm F}^{A=1} + \mathcal{L}_{\rm F}^{A=0} +  \alpha\cdot  {\rm IPM}_{G}(p_\Phi^{A=1},p_\Phi^{A=0}) \,.
		\end{align}
	\end{corollary}
	The integral probability metric (IPM) \citep{muller1997integral, sriperumbudur2012empirical} measures the distance between two probability distributions $p$ and $q$ defined over $M$, \textit{i.e.}, the latent space of $R$.
	Formally, \newline
	${\rm IPM}_G (p, q) \triangleq \sup_{g \in G} |\int_{M} g(m) \left(p(m) -q(m) \right) dm|$,  where $g:m \rightarrow \mathbb{R}$, represents a class of real-valued bounded measurable functions on $M$ \citep{shalit2017estimating}.
	Therefore, model functions $\{h_a(\cdot), \Phi(\cdot)\}$ can be learned by minimizing the upper bound in \eqref{eq:upper_bound} consisting of 
	$(i)$ only \emph{factual} losses under both treatment assignments and 
	$(ii)$ an IPM regularizer enforcing latent distributional equivalence between the treatment groups.
	Note that if the data originates from a RCT it follows (by construction) that ${\rm IPM}_{G}(p_\Phi^{A=1},p_\Phi^{A=0}) = 0$.
	
	\subsection{Accounting for censoring bias}
	%
	Below we formulate an approach for estimating functions $h_A(\cdot)$ and $\nu_A(\cdot)$ for synthesizing (sampling) non-censored $t_a  \sim p_{h, \Phi} (T_A |X=x)$ and censored $c_a  \sim p_{\nu, \Phi} (C_A |X=x)$ times, respectively.
	While some parametric assumptions for $p_{h,\Phi}(T_A |X=x)$ yield easy-to-evaluate closed forms for $S_{h,\Phi}(t_a|x)$ that can be used as likelihood for censored observations, they are restrictive, and have been shown to generate unrealistic high variance samples \citep{chapfuwa2018adversarial}.
	So motivated, we seek a nonparametric likelihood-based approach that can model a flexible family of distributions, with an easy-to-sample approach for event times $t_a \sim p_{h,\Phi}(T_a|X=x)$.
	%
	We model the event time generation process with a source of randomness, $p({\epsilon})$, {\em e.g.} Gaussian or uniform, which is obtained from a neural-network-based nonlinear transformation. In the experiments we use a \emph{planar flow} formulation parameterized by $\{U_h, W_h, b_h\}$ \citep{rezende2015variational}, however, other specifications can also be used. Note that \cite{miscouridou2018deep} has previously leveraged normalizing flows for survival analysis, however, our approach is very different in that it focuses on $i$) formulating a counterfactual survival analysis framework that accounts for \emph{informative or non-informative} censoring mechanisms and confounding, and $ii$) modeling event times as a continuous variable instead of discretizing them.
	Specifically, we transform the source of randomness, $\epsilon$, using a single layer specification as follows
	%
	\begin{align}
	\label{eq:epsilonv}
	\begin{aligned}
	\tilde{\epsilon}_h  & = {\epsilon} + {U}_h \tanh(W_h \epsilon + b_h)\,, \quad \epsilon \sim  {\rm Uniform}(0,1) \,, \\
	t_a  &= h_A({r}, \tilde{\epsilon}_h)\,, \quad {r} = \Phi(x) 
	\end{aligned}
	\end{align}
	where $\{U_h, W_h\} \in \mathbb{R}^{d \times d}$, $\{ b_h, \epsilon \} \in \mathbb{R}^ d$, $d$ is the dimensionality of the planar flow; each component of $\epsilon$ is drawn independently from ${\rm Uniform}(0,1)$, and $\tilde{\epsilon}_h$ may be viewed as a skip connection with stochasticity in $\epsilon$.
	Further, $h_A(r,\tilde{\epsilon}_h)$ and $\Phi(x)$ are time-to-event generative and encoding functions, respectively, parameterized as neural networks.
	For simplicity, the dimensions of ${r}$ and $\epsilon$ are set to $d$, however, they can be set independently if desired.
	In practice, we are interested in generating realistic event-time samples; therefore, we account for both censored and non-censored observations by adopting the objective from \citet{chapfuwa2018adversarial}, formulated as 
	\begin{align}\label{eq:nfr_loss}
	\begin{aligned}
	\mathcal{L}_{\rm F}^{\rm CSA} & \triangleq  \mathbb{E}_{(y,\delta,x,a) \sim p(Y,\delta,X,A), \epsilon \sim p(\epsilon)} \left[ \delta \cdot  \left(|y-t_a| \right) \right. \\
	&  + \left. (1-\delta) \cdot  \left(\max(0, y-t_a)\right) \right]\,,
	\end{aligned}
	\end{align}
	where the first term encourages sampled event times $t_a$ to be close to $y$, the ground truth for observed events, \textit{i.e.}, $\delta=1$, while penalizing $t_a$ for being smaller than the censoring time when $\delta=0$.
	Further, the expectation is taken over samples (a minibatch) from empirical distribution $p(Y,\delta,X,A)$.
	
	\paragraph{Informative censoring}
	We model informative censoring similar to \eqref{eq:nfr_loss} but mirroring the censoring indicators to encourage accurate censoring time samples $c_a$ for $\delta=0$, while penalizing $c_a$ for being smaller than $y$ for $\delta=1$ (observed events).
	Specifically, we set an independent source of randomness like in \eqref{eq:epsilonv} but parameterized by $\{{U}_{\nu}, {W}_{\nu}, b_{\nu}\}$ and censoring generative functions $\nu_A ({r}, \tilde{\epsilon}_\nu)$, parameterized as neural networks, where $c_a \sim p_{\nu, \Phi} (C_A|X=x)$ formulated as
	\begin{align}\label{eq:censoring_reg}
	\begin{aligned}
	\ell_{c}( \nu, \Phi) &= \mathbb{E}_{(y,\delta,x,a) \sim p(y,\delta, X, A), \epsilonv \sim p(\epsilon)}  \left[ (1-\delta) \cdot \left(|y-c_a| \right)  \right. \\
	&  + \left.  \delta \cdot \left(\max(0, y-c_a)\right) \right] \,.
	\end{aligned}
	\end{align}
	Further, we introduce an additional time-order-consistency loss that enforces the correct order of the observed time relative to the censoring indicator, \textit{i.e.}, $c_a < t_a$ if $\delta=0$ and $t_a < c_a$ if $\delta=1$, thus 
	\begin{align}\label{eq:event_reg}
	\begin{aligned}
	\ell_{\rm TC}( h, \nu, \Phi) &= \mathbb{E}_{(\delta,x,a)\sim p(\delta,X,A),\epsilon \sim p(\epsilon)} 
	\left[ \delta \cdot \left(\max(0,t_a-c_a) \right) \right. \\
	&+ \left. (1-\delta) \cdot \left( \max(0,c_a-t_a)\right) \right] \,.
	\end{aligned}
	\end{align}
	Note that $\ell_{\rm TC}( h, \nu, \Phi)$ does not depend on the observed event times but only on the censoring indicators.
	Finally, we write the consolidated CSA loss for informative censoring (CSA-INFO) by aggregating \eqref{eq:nfr_loss}, \eqref{eq:censoring_reg} and \eqref{eq:event_reg} as
	\begin{align}\label{eq:nfr_info_loss}
	\mathcal{L}_{\rm F}^{\rm CSA-INFO} & \triangleq  \mathcal{L}_{\rm F}^{\rm CSA}  + \ell_{\rm c} +  \ell_{\rm TC}\,.
	\end{align}
	
	\begin{table*}
		\centering
		\caption{Performance comparisons on {\sc actg-Synthetic} data, with 95\% ${\rm HR} (t)$ confidence interval. The ground truth, test set, hazard ratio is $\rm{HR(t)} = 0.52_{(0.39, 0.71)}$.
		}
		\label{tb:actg-sythentic}
		\vspace{-2mm}
		\adjustbox{max width=1.0\textwidth}{
			\begin{tabular}{l|rrr|rrr}
				\multirow{2}{*}{\textbf{Method} }&
				\multicolumn{3}{c|}{\textbf{Causal metrics}}&
				\multicolumn{3}{c}{\textbf{Factual metrics}} \\
				& $\epsilon_{\rm PEHE}$ &$\epsilon_{\rm ATE}$ &  ${\rm HR}(t)$  & C-Index (A=0, A=1)& Mean COV&  C-Slope  (A=0, A=1) \\
				\toprule
				CoxPH-Uniform  & NA &   NA & $0.97_{(0.86, 1.09)}$&  NA&  NA& NA \\
				CoxPH-IPW  & NA &   NA & $0.48_{(0.03, 7.21)}$&  NA&  NA& NA \\
				CoxPH-OW   & NA &   NA & $0.60_{(0.53, 0.68)}$&  NA&  NA& NA \\
				Surv-BART &  352.07  & 77.89  & $0.0 (0.0, 0.0)$ & (0.706,  0.686) &  0.001 & (0.398,  $\infty$) \\
				AFT-Weibull  & 367.92& 133.93 &$ 0.47_{(0.47, 0.47)}$ & (0.21, 0.267) &6.209 & (0.707, 0.729)  \\
				AFT-log-Normal   &377.76 &  157.64&$ 0.47_{(0.47, 0.47)}$ & (0.675, 0.556) &6.971 & (0.707, 0.729)  \\
				SR   & 369.47 & 88.55 &$0.38_{(0.33, 0.65)}$ & (0.791, 0.744) &0& (0.985, 1.027)  \\
				\hdashline
				CSA (proposed) & 358.72 & \textbf{0.8} &$0.45_{(0.39, 0.65)}$ & (0.787, 0.767) &0.131 & (0.985, 1.026)  \\
				CSA-INFO (proposed) &\textbf{344.3} & 31.19  &$\textbf{0.53}_{(0.41, 0.67)}$ & (0.78, 0.764) & 0.13& (0.999, 1.029)  \\
			\end{tabular}
		}
	\end{table*}
	\subsection{Learning}
	Model functions $\{h_A(\cdot), \Phi(\cdot), \nu_A(\cdot)\}$ are learned by minimizing the bound \eqref{eq:upper_bound}, via stochastic gradient descent on minibatches from $\CD$, with $\mathcal{L}_{\rm F}^{\rm CSA}$ for non-informative censoring and $\mathcal{L}_{\rm F}^{\rm CSA-INFO}$ for informative censoring.
	Further, for the ${\rm IPM}$ regularization loss in \eqref{eq:upper_bound}, we optimize the dual formulation of the \emph{Wasserstein distance}, via the regularized \emph{optimal transport} \citep{villani2008optimal, cuturi2013sinkhorn}.
	Consequently, we only require $\alpha^{-1}\zeta_{h,\Phi}(x,a)$ to be $1$-Lipschitz \citep{shalit2017estimating} and $\alpha$ is selected by grid search on the validation set using \textit{only} factual data (details below).
	
	\section{Metrics}
	
	We propose a comprehensive evaluation approach that accounts for both factual and causal metrics.
	Factual survival outcome predictions are evaluated according to standard survival metrics that measure diverse performance characteristics, such as concordance index (C-Index) \citep{harrell1984regression}, mean coefficient of variation (COV) and calibration slope (C-slope) \citep{chapfuwa2020survival}.
	See the SM for more details on these metrics.
	For causal metrics, defined below, we introduce a nonparametric hazard ratio (HR) between treatment outcomes, and adopt the conventional precision in estimation of heterogeneous effect (PEHE) and average treatment effect (ATE) performance metrics \citep{hill2011bayesian}.
	Note that PEHE and ATE require ground truth counterfactual event times, which is only possible for (semi-)synthetic data.
	For HR, we compare our findings with those independently reported in the literature from gold-standard RCT data.

	\begin{table*}
		\centering
		\caption{Performance comparisons on {\sc Framingham} data, with 95\% ${\rm HR}(t)$  confidence interval. Test set NN assignment of  $y_{\rm{CF}}$ and $\delta_{\rm{CF}}$ yields biased ${\rm HR(t)}=1.23_{(1.17, 1.25)}$, while previous large scale longitudinal RCT studies estimated ${\rm HR (t)}= 0.75_{(0.64,  0.88)}$ \citep{yusuf2016cholesterol}.}
		\label{tb:framingham}
		\vspace{-2mm}
		\adjustbox{max width=0.95\textwidth}{
			\begin{tabular}{l|r|rrr}
				\multirow{2}{*}{\textbf{Method} }&
				\multicolumn{1}{c|}{\textbf{Causal metric}}&
				\multicolumn{3}{c}{\textbf{Factual metrics}} \\
				& ${\rm HR}(t)$   & C-Index (A=0, A=1)& Mean COV&  C-Slope  (A=0, A=1) \\
				\toprule
				CoxPH-Uniform  & $1.69_{(1.38,2.07)}$&  NA&  NA& NA \\
				CoxPH-IPW  &  $1.09_{(0.76,1.57)}$&  NA&  NA& NA \\
				CoxPH-OW   & $0.88_{(0.73, 1.08)}$ &  NA&  NA& NA \\
				Surv-BART &  $14.99 _{(14.9, 14.9e8)}$  & (0.629,  0.630) & 0.003  & (0.232,  0.084) \\
				AFT-Weibull   & $1.09_{(1.09, 1.09)}$ & (0.734, 0.395) & 8.609& (0.857, 0.89)  \\
				AFT-log-Normal & $1.55_{(1.46, 1.55)}$ & (0.68, 0.56) & 10.415& (0.979, 0.732)  \\
				SR     & $0.58_{(0.53, 0.71)}$ & (0.601, 0.57) &0 & (0.491, 0.63)  \\
				\hdashline
				CSA (proposed) & $1.04_{(1.00, 1.09)}$ & (0.763, 0.728) &0.161 & (0.891, 0.81)  \\
				CSA-INFO (proposed) & $ \textbf{0.81}_{(0.77, 0.83)}$ & (0.752, 0.651) &0.156 & (0.907, 0.881)  \\
			\end{tabular}
		}
	\end{table*}
	
	\paragraph{Nonparametric Hazard Ratio}
	In medical settings, the population hazard ratio ${\rm HR}(t)$ between treatment groups is considered informative thus has been widely used in drug development and RCTs \citep{yusuf2016cholesterol, mihaylova2012effects}.
	For example, ${\rm HR} (t) < 1$, $> 1$, or $\approx 1$ indicate \emph{population} positive, negative and neutral treatment effects at time $t$, respectively.
	Moreover, ${\rm HR}(t)$ naturally accounts for both censored and non-censored outcomes.
	Standard approaches for computing ${\rm HR} (t)$ rely on the restrictive proportional hazard assumption from  CoxPH \citep{cox1972regression}, which is constituted as a semi-parametric linear model $\lambda(t|a) = \lambda_{\rm b}(t) \exp(a \beta)$.
	However, the constant covariate (time independent) effect is often violated in practice (see Figure~\ref{fig:subpop_hazard}).
	For CoxPH, the \emph{marginal} HR between treatment and control can be obtained from regression coefficient $\beta$ learned via maximum likelihood without the need for specifying the baseline hazard $\lambda_{\rm b}(t)$:
	\begin{align}
	\label{eq:cox-hr}
	{\rm HR}_{\rm CoxPH} (t) &= \frac{\lambda (t|a=1)}{\lambda (t|a=0)} = \exp(\beta)\,.
	\end{align}
	%
	So motivated, we propose a nonparametric, model-free approach for computing ${\rm HR} (t)$, in which we do not assume a parametric form for the event time distribution or the proportional hazard assumption from CoxPH. This approach only relies on samples from the conditional event time density functions, $f(t_1|x)$ and $f(t_0|x)$, via $t_a=h_A(\cdot)$ from \eqref{eq:epsilonv}.
	%
	\begin{definition}{We define the nonparametric marginal Hazard Ratio and its approximation, ${\rm \hat{HR}}(t)$, as}
		%
		\begin{align}
		\label{eq:nonpar-hr}
		\begin{aligned}
		{\rm HR} (t) &= \frac{\lambda_1(t)}{\lambda_0(t)} =  \frac{S_0(t)} {S_1(t)} \cdot  \frac{S_1^{\prime} (t)} { S_0^{\prime} (t)} \,, \\ 
		{\rm \hat{HR}}(t)  &=  \frac{\hat{S}_0^{\rm PKM}(t)} {\hat{S}_1^{\rm PKM}(t)} \cdot  \frac{m_1 (t)} { m_0(t)}\,,
		\end{aligned}
		\end{align}
	\end{definition}
	
	where for ${\rm HR} (t)$ we leveraged \eqref{eq:hazard} to obtain \eqref{eq:nonpar-hr} and $S^{\prime}(t) \triangleq d S(t) /dt$.
	The nonparametric assumption for $S(t)$ makes the computation of $S^{\prime}(t)$  challenging.
	Provided that $S(t)$ is a monotonically decreasing function, for simplicity, we fit a linear function $S(t) = m \cdot t + c$, and set $S^{\prime}(t) \approx m$.
	Note that the linear model is \emph{only} used for estimating $S^{\prime}(t)$ from the nonparametric estimation of $S(t)$.
	Bias from $S^{\prime}(t)$ can be reduced by considering more complex function approximations for $S(t)$, \textit{e.g.}, polynomial or spline.
	For the nonparametric estimation of $S(t)$ we leverage the \emph{model-free} population point-estimate-based nonparametric Kaplan-Meier \citep{kaplan1958nonparametric} estimator of the survival function $\hat{S}^{\rm PKM} (t)$ in \citep{chapfuwa2020survival} to marginalize both \emph{factual} and \emph{counterfactual} predictions given covariates $x$.
	The approximated hazard ratio, ${\rm \hat{HR}}(t)$, is thus obtained by combining the approximations $\hat{S}^{\rm PKM}_a (t)$ and $m_a$.
	A similar formulation for the conditional, ${\rm \hat{HR}} (t|x)$, can also be derived.
	See the SM for full details on the evaluation of ${\rm \hat{HR}}(t)$ and ${\rm \hat{HR}} (t|x)$.
	Note that for some AFT- or CoxPH-based parametric formulations, ${\rm HR} (t|x)$, can be readily evaluated because $f(t_a|x)$ and $S(t_a|x)$ are available in closed form.
	
	In the experiments, we will use ${\rm HR}(t)$ to compare different approaches against results reported in RCTs (see Tables \ref{tb:actg-sythentic} and \ref{tb:framingham}).
	Further, we will use ${\rm HR}(t|x)$ to illustrate {\em stratified} treatment effects (see Figure 2).
	Note that though a neural-network-based survival recommender system \citep{katzman2018deepsurv} has been previously used to estimate ${\rm HR} (t|x)$, their approach does not account for confounding or informative censoring thus it is susceptible to bias.
	
	
	\paragraph{Precision in Estimation of Heterogeneous Effect (PEHE)} A general \emph{individualized} estimation error is formulated as 
	$$\epsilon_{\rm PEHE} = \sqrt{\mathbb{E}_{X} [ ({\rm ITE}(x) -\hat{\rm ITE}(x) )^2]},$$
	where ${\rm ITE} (x)$ is the ground truth, $\hat{\rm ITE} (x) = \mathbb{E}_{T}\left[\gamma\left( T_1 \right) - \gamma\left( T_0 \right) | X = x\right]$ and $\gamma(\cdot)$ is a deterministic transformation.
	In our experiments, $\gamma(\cdot)$ is the average over samples from $t_a  \sim p_{h, \Phi} (T_A |X=x)$.
	Alternative estimands, \textit{e.g.}, thresholding survival times $\gamma(T_A) = I \{ T_A >\tau \}$, can also be considered as described in Section~\ref{sec:estimands}.
	
	\paragraph{Average Treatment Effect (ATE)} The \emph{population} treatment effect estimation error is defined as $$\epsilon_{\rm ATE} = |{\rm ATE} - \hat{\rm ATE} |,$$
	where ${\rm ATE} =\mathbb{E}_{X}[ { \rm ITE} (x)]$ (ground truth) and $\hat{\rm ATE} =\mathbb{E}_{X}[ \hat{\rm ITE} (x)]$.
	
	Note that both PEHE and ATE require ground truth (population and individual) treatment effects to be available, which is only possible in synthetic and semi-synthetic data (benchmarking) scenarios.

	\section{Experiments}
	

	We describe the baselines and datasets that will be used to evaluate the proposed counterfactual survival analysis methods (CSA and CSA-INFO). Detailed architecture information of the proposed methods (CSA and CSA-INFO) and baselines (AFT-log-Normal, AFT-Weibull, Semi-supervised Regression(SR)) are provided in the SM. Pytorch code to replicate experiments can be found at \textcolor{blue}{\url{https://github.com/paidamoyo/counterfactual_survival_analysis}}.
	Throughout the experiments, we use the standard ${\rm HR}(t)$ for CoxPH based methods in \eqref{eq:cox-hr} and \eqref{eq:nonpar-hr} for all others.
	The bound in \eqref{eq:upper_bound} is sensitive to $\alpha$, thus we propose approximating \textit{proxy} counterfactual outcomes $\{Y_{\rm CF}, \delta_{\rm CF} \}$ for the validation set, according to the covariate Euclidean nearest-neighbour (NN) from the training set.
	We select the $\alpha$ that minimizes the validation loss $\mathcal{L} = \mathcal{L}_{\rm F} + \mathcal{L}_{\rm CF}$ from the set $(0, 0.1, 1, 10, 100)$.
	
	\paragraph{Baselines}
	We consider the following competitive baseline approaches: 
	$(i)$ \emph{propensity} weighted CoxPH \citep{schemper2009estimation, buchanan2014worth, rosenbaum1983central}; 
	$(ii)$ ${\rm IPM}$ \eqref{eq:upper_bound} regularized AFT (log-Normal and Weibull) models; 
	$(iii)$ an ${\rm IPM}$ \eqref{eq:upper_bound} regularized \textit{deterministic} semi-supervised regression (SR) model with accuracy objective from \citep{chapfuwa2018adversarial}, as a contrast for the proposed stochastic predictors (CSA and CSA-INFO); and $(iv)$ survival Bayesian additive regression trees (Surv-BART) \citep{sparapani2016nonparametric}.
	For CoxPH, we consider three normalized weighting schemes: 
	$(i)$ inverse probability weighting (IPW) \citep{horvitz1952generalization, cao2009improving}, where ${\rm IPW}_i  = \frac{a_i}{\hat{e}_i} + \frac{1-a_i}{1-\hat{e}_i}$; 
	$ii)$ overlapping weights (OW) \citep{crump2006moving, li2018balancing}, where ${\rm OW}_i = a_i \cdot (1-\hat{e}_i) + (1-a_i) \cdot \hat{e}_i$; and 
	$iii)$ the standard RCT uniform assumption.
	A simple linear logistic model $\hat{e}_i = \sigma(x_i; w)$, is used as an approximation, $\hat{e}_i$, to the unknown propensity score $P(A=1|X=x)$.
	See the SM for more details of the baselines.
	
	\begin{figure*}[!htbp]
		\begin{subfigure}{.33\textwidth}
			\centering
			\includegraphics[width=.86\linewidth]{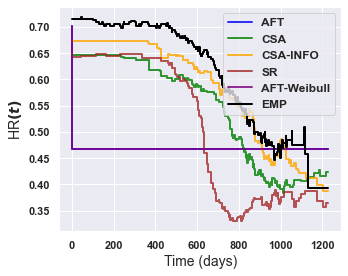}  
			\caption{ {\sc actg-synthentic} ${\rm HR}(t)$}
			\label{fig:pop_hazard}
		\end{subfigure}
		\begin{subfigure}{.31\textwidth}
			\centering
			\includegraphics[width=.94\linewidth]{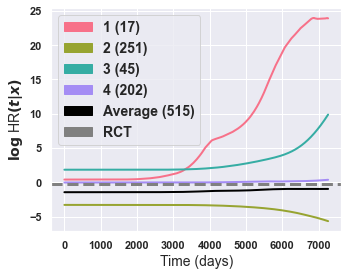}
			\caption{ {\sc Framingham} $ \log {\rm HR}(t|x)$ }
			\label{fig:subpop_hazard}
		\end{subfigure}
		\begin{subfigure}{.33\textwidth}
			\centering
			\includegraphics[width=.88\linewidth]{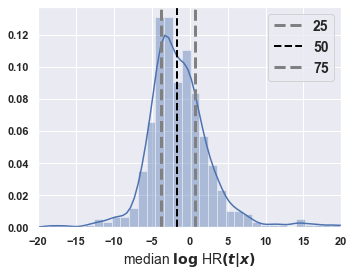}
			\caption{{\sc Framingham} $\log {\rm HR}(t|x)$ pdf}
			\label{fig:subpob_pdf}
		\end{subfigure}
		\vspace{-2mm}
		\caption{(a) Inferred population ${\rm HR}(t)$ compared against ground truth (EMP) on {\sc actg-Synthetic} data. CSA-INFO-based (b) cluster-specific average $\log {\rm HR} (t|x)$ curves and (c) estimated density of median $\log {\rm HR} (t|x)$ values on the test set of the {\sc Framingham} dataset. Clusters assignment were obtained via hierarchical clustering of individualized $\log {\rm HR}(t|x)$ traces.}
		\label{fig:hazards}
	\end{figure*}
	
	\paragraph{Datasets}
	We consider the following datasets summarized in Table~\ref{tb:data}:
	$(i)$ {\sc Framingham}, is an EHR-based longitudinal cardiovascular cohort study that we use to evaluate the effect of statins on future coronary heart disease outcomes \citep{benjamin1994independent}; 
	$(ii)$ {\sc actg}, is a longitudinal RCT study comparing monotherapy with Zidovudine or Didanosine with combination therapy in HIV patients \citep{hammer1996trial}; and 
	$(iii)$ {\sc actg-Synthetic}, is a semi-synthetic dataset based on {\sc actg} covariates.
	We simulate potential outcomes according to a Gompertz-Cox distribution \citep{bender2005generating} with selection bias from a simple logistic model for $P(A=1| X=x)$ and AFT-based censoring mechanism.
	The generative process is detailed in the SM.
	Table \ref{tb:data} summarizes the datasets according to 
	$(i)$ covariates of size $p$; 
	$(ii)$ proportion of non-censored events, treated units, and missing entries in the $N \times p$ covariate matrix; and 
	$(iii)$ time range $t_{\rm max}$ for both censored and non-censored events.
	Missing entries are imputed with the median or mode if continuous or categorical, respectively.
	
	\begin{table}[!t]
		\caption{Summary statistics of the datasets.}
		\label{tb:data}
		\vspace{-2mm}
		\resizebox{0.45\textwidth}{!}{
			\begin{tabular}{lrrr}
				& {\sc Framingham}  & {\sc actg} &  {\sc actg-synthetic}  \\
				\toprule
				Events (\%) &  26.0  & 26.9 & 48.9\\
				Treatment (\%) & 10.4  & 49.5 & 55.9\\
				$N$ &  3,435  & 1,054 & 2,139\\
				$p$  & 32  & 23  & 23\\
				Missing (\%) &  0.23  & 1.41  & 1.38\\
				$t_{\rm max}$ (days) & 7,279 & 1,231 & 1,313 \\
				\bottomrule
			\end{tabular}
		}
			\vspace{-3mm}
	\end{table}
	
	\paragraph{Quantitative Results}
	Experimental results for two data-sets in Tables \ref{tb:actg-sythentic} and \ref{tb:framingham}, illustrate that AFT-based methods have high variance, inferior in calibration and C-Index than accuracy-based methods (SR, CSA, CSA-INFO).
	Surv-BART is the least calibrated but low variance method.
	CSA-INFO and CSA outperform all methods across all factual metrics, whereas CSA-INFO is better calibrated, has low variance but slightly lower C-Index than CSA.
	Note that we fit CoxPH using the entire dataset; since it does not support counterfactual inference, we do not present factual metrics.
	By properly adjusting for both informative censoring and selection bias, CSA-INFO significantly outperforms all methods in treatment effect estimation according to ${\rm HR} (t)$ and $\epsilonv_{\rm PEHE}$, across non-RCT datasets, while remaining comparable to AFT-Weibull on the RCT dataset (see the SM).
	Further, RCT-based results on {\sc actg} data in the SM illustrate comparable ${\rm HR} (t)$ across all models except for AFT-log-Normal and Surv-BART, which overestimate, and SR, which underestimates risk.
	For non-RCT datasets ({\sc actg-Synthentic} and {\sc framingham}), CoxPH-OW has a clear advantage over all CoxPH based methods, mostly credited to the well-behaved bounded propensity weights $\in [0, 1]$.
	Interestingly, the {\sc Framingham} observational data exhibits a common paradox, where without proper adjustment of selection and censoring bias, naive approaches would result in a counter-intuitive treatment effect from statins.
	However, there is severe \emph{confounding} from covariates such as age, BMI, diabetes, CAD, PAD, MI, stroke, \textit{etc}., that influence both treatment likelihood and survival time.
	Table \ref{tb:framingham}, demonstrates that CSA-INFO is clearly the best performing approach. Specifically, its ${\rm HR} (t)$, reverses the biased observational treatment effect, to demonstrate positive treatment from statins, which is consistent with prior large RCT longitudinal findings \citep{yusuf2016cholesterol}. Consequently, our experiments are comprehensive and we are confident that the CSA-INFO performance benefits are attributed to $(i)$ accounting for informative censoring bias; $(ii)$  accounting for selection bias (optimal IPM regularizer with  $\alpha > 0$); and $(iii)$  {\em flexible and non-parametric} generative modeling of event times from the stochastic planar flow.

	\paragraph{Qualitative Results}
	Figure \ref{fig:pop_hazard} demonstrates that CSA-INFO matches the ground truth population hazard, ${\rm HR} (t)$, better than alternative methods on {\sc actg-Synthetic} data.
	See the SM for {\sc actg} and {\sc framingham}. 
	Figure~\ref{fig:subpop_hazard} shows sub-population log hazard ratios for four patient clusters obtained via hierarchical clustering on the individual log hazard ratios, $\log {\rm HR} (t|x)$, of the test set of {\sc Framingham} data.
	Interestingly, these clusters stratify treatment effects into: positive (2), negative (1 and 3), and neutral (4) sub-populations.
	Moreover, the estimated density of median $\log {\rm HR}(t|x)$ values in Figure~\ref{fig:subpob_pdf} illustrates that nearly $70\%$ of the testing set individuals have $\log {\rm HR}(t|x)<0$, thus may benefit from taking statins.
	Further, we isolated the extreme top and bottom quantiles, ${\rm HR}(t|x)<0.024$ and ${\rm HR}(t|x)>1.916$, respectively, of the median $\log {\rm HR}(t|x)$ values for the test set of {\sc Framingham}, as shown in Figure~\ref{fig:subpob_pdf}.
	After comparing their covariates, we found that individuals with the following characteristics may benefit from taking statins: young, male, diabetic, without prior history (CAD, PAD, stroke or MI), high BMI, cholesterol, triglycerides, fasting glucose, and low high-density lipoprotein. Note that individuals with contrasting covariates experience may not benefit from taking statins. 
	There seem to be consensus that diabetics and high-cholesterol patients benefit from statins \citep{cheung2004meta, wilt2004effectiveness}.
	See SM for additional results.
	
	
	\section{Conclusions}
	%
	We have proposed a unified counterfactual inference framework for survival analysis.
	Our approach adjusts for bias from two sources, namely, \emph{confounding} (covariates influence both the treatment assignment and the outcome) 
	and \emph{censoring} (informative or non-informative).
	Relative to competitive alternatives, we demonstrate superior performance for both survival-outcome prediction and treatment-effect estimation, across three diverse datasets, including a semi-synthetic dataset which we introduce.
	Moreover, we formulate a model-free nonparametric hazard ratio metric for comparing treatment effects or leveraging prior randomized real-world experiments in longitudinal studies. We demonstrate that the proposed model-free hazard-ratio estimator can be used to identify or stratify heterogeneous treatment effects. Finally, this work will serve as an important baseline for future work in real-world counterfactual survival analysis. In future work,  we plan to understand the sensitivity of our estimates to unobserved confounding \cite{cornfield1959smoking} and the effect of both censoring bias and selection bias on causal identifiability.

	\section*{Acknowledgments}
	The authors would like to thank the anonymous reviewers for their insightful comments.
	This work was supported by NIH/NIBIB R01-EB025020 and NIH/NINDS 1R61NS120246-01.
	
	\bibliographystyle{ACM-Reference-Format}
	\bibliography{causal_sa}

\include{causal_sa_SM}

\end{document}

%% file: causal_sa_SM.tex
\setcopyright{acmcopyright}
\copyrightyear{2021}
\acmYear{2021}
\acmDOI{10.1145/1122445.1122456}

	\appendix
	\title{ Supplementary Material for ``Enabling Counterfactual Survival Analysis with Balanced Representations"}
	
	\author{Anonymous Authors}
	\affiliation{
		\institution{Anonymous Institution}
		\city{Anonymous City}
		\country{Anonymous Country}
	}

\settopmatter{printacmref=false}
\setcopyright{none}
\renewcommand\footnotetextcopyrightpermission[1]{}
\pagestyle{plain}

\setcopyright{none}
\makeatletter
\renewcommand\@formatdoi[1]{\ignorespaces}
\makeatother

	\renewcommand{\shortauthors}{Anonymous Authors, et al.}


	\section{ General  log-likelihood}
	
	The general  likelihood-based loss hypothesis that accounts for informative censoring is formulated as: 
	\begin{align}\label{eq:gen_lik}
	- \ell_{h, \Phi, \nu} (t_a,  c_a, y, \delta) & = \log  p_{h,\Phi, \nu}(T_A, C_A|X=x) \\
	\label{eq:cond_ind}
	& = \log p_{h,\Phi}(T_A|X=x)  + \log  p_{\nu,\Phi}(C_A| X=x) \, ,
	\end{align}
	where \eqref{eq:cond_ind} follows from  the conditional independence  (informative censoring) assumption $T \bigCI C | X, A$.
	For some parametric formulations of event  $ p_{h, \Phi} (T_A|X=x)$ and censoring  $ p_{\nu, \Phi} (C_A |X=x)$ time distributions, \textit{e.g.}, exponential, Weibull, log-Normal, \textit{etc}., then $-\ell_{h, \Phi, \nu} ( t_a, c_a,y, \delta)$ is the closed-form  log-likelihood, where:
	\begin{align}
	\log p_{h,\Phi}(T_A|X=x) & \triangleq \delta \cdot \log  f_{h,\Phi}(t_a|x )+  (1-\delta) \cdot  \log S_{h,\Phi}(t_a|x) , \\
	\log p_{\nu,\Phi}(C_A|X=x) & \triangleq  (1-\delta) \cdot \log e_{\nu,\Phi}(c_a|x) + \delta \cdot  \log G_{\nu,\Phi}(c_a|x) ,
	\end{align}
	where  $\{ S_{h,\Phi} (\cdot), G_{\nu, \Phi} (\cdot) \}$ and $\{ f_{h,\Phi} (\cdot) , e_{\nu, \Phi} (\cdot) \}$ are survival and density functions respectively. 
	
	\section{Metrics}
	\subsection{Estimands of Interest}
	Several common estimands of interest include \citep{zhao2012utilizing, trinquart2016comparison}:
	\begin{itemize}[topsep=0pt,itemsep=0pt,parsep=0pt,partopsep=0pt,leftmargin=10pt]
		\item {\em Difference in expected lifetime}: \newline
		 $\textup{ITE}(x)=\int_{0}^{t_{\max}}\{S_{1}( t|x)-S_{0}(t|x)\}\textup{d}t=\mathbb{E}\{T_1 -T_0|X =x\}$.
		\item {\em Difference in survival function}: $\textup{ITE}(t, x)=S_{1}(t |x)-S_{0}(t |x)$.
		\item { \em Hazard ratio}: $\textup{ITE}(t, x)=\lambda_{1}(t |x)/\lambda_{0}(t|x)$.
	\end{itemize}
	In our experiments, we consider both the hazard ratio and difference in expected lifetime.  The difference of expected lifetime is expressed in terms of both survival functions and expectations:
	\begin{align}
	\mathbb{E}[T|X=x] &= \int_{-\infty} ^{\infty} t f(t|x) dt \notag \\
	\label{eq:expectation_prop}
	&=  \int_{0} ^{\infty} \left(1-F(t|x)\right) dt - \int_{-\infty} ^{0} F(t|x) dt\\
	\label{eq:lifetime}
	& = \int_{0} ^{t_{\max} } S(t|x) dt\,,
	\end{align}
	where \eqref{eq:expectation_prop} follows from standard properties of expectations and \eqref{eq:lifetime} from $1-F(t|x) = S(t|x)$ and $\int_{-\infty} ^{0} F(t|x) dt = 0$.
	Below we formulate an approach for estimating the individualized and population hazard ratio. 
	
	\subsection{Nonparametric Hazard Ratio}
%
	To estimate the proposed  \textit{nonparametric hazard ratio}  ${\rm HR} (t)$ in   \eqref{eq:nonpar-hr} we leveraged \eqref{eq:hazard} and $S^{\prime}(t) \triangleq d S(t) /dt$.  For the estimator ${\rm \hat{HR}}(t) $, provided that $S(t)$ is a monotonically decreasing function, for simplicity, we fit a linear function $S(t) = m \cdot t + c$ and set $S^{\prime}(t) \approx m$. Further, we leverage $\hat{S}^{\rm PKM} (t)$ in \citep{chapfuwa2020survival},  defined as the model-free population point-estimate-based nonparametric Kaplan-Meier \citep{kaplan1958nonparametric} estimator. We denote $J$ distinct and ordered  observed event times (censored and non-censored) by the set  $\mathcal{T} = \{ t_j |  t_j > t_{j-1} > \ldots > t_0\}$ from  $N$ realizations of $Y$. Formally,  the \emph{population} survival  $\hat{S}^{\rm PKM} _A(t)$  is recursively formulated as 
	\begin{align}\label{eq:pop_pkm}
	\hat{S}^{\rm PKM}_A(t_j ) & = \left(1- \frac{\sum_{n:\delta_n=1} \mathbb{I}\left(t_{j-1} \le \gamma (T_A^{(n)}) <  t_{j} \right)  }{ N - \sum_{n=1}^{N}  \mathbb{I} \left( \gamma (T_A^{(n)}) < t_{j-1} \right)  } \right) \hat{S}^{\rm PKM}_A(t_{j-1})\,,
	\end{align}
	where $\hat{S}_A^{\rm PKM}(t_0) = 1$, and $\mathbb{I}(b)$ represent an  indicator function such that $\mathbb{I}(b)=1$ if $b$ holds or $\mathbb{I}(b)=0$ otherwise.  Further, $\gamma(\cdot)$ is a deterministic transformation for summarizing $T_A$, in our experiments, $\gamma(\cdot) = {\rm median(\cdot)}$, computed over samples from $t_a  \sim p_{h, \Phi} (T_A |X=x)$. Note from \eqref{eq:pop_pkm},  we marginalize both \emph{factual} and \emph{counterfactual} predictions given covariates $x$.
	
	\begin{table*}
		\centering
		\caption{Performance comparisons on {\sc actg} data, with 95\% ${\rm HR}(t)$  confidence interval. Test set NN assignment of  $y_{\rm{CF}}$ and $\delta_{\rm{CF}}$ yields unbiased ground truth estimator ${\rm HR}(t)=0.54_{(0.51, 0.61)}$, since study is a RCT.}
		\label{tb:actg}
		\vspace{1mm}
		\adjustbox{max width=0.85\textwidth}{
			\begin{tabular}{l|r|rrr}
				\multirow{2}{*}{\textbf{Method} }&
				\multicolumn{1}{c|}{\textbf{Causal metric}}&
				\multicolumn{3}{c}{\textbf{Factual metrics}} \\
				& ${\rm HR}(t)$   & C-Index (A=0, A=1)& Mean COV&  C-Slope  (A=0, A=1) \\
				\toprule
				CoxPH-Uniform  & $0.49_{(0.38,0.64)}$&  NA&  NA& NA \\
				CoxPH-IPW   & $0.49_{(0.36,0.68)}$&  NA&  NA& NA \\
				CoxPH-OW   &  $0.49_{(0.36,0.68)}$&  NA&  NA& NA \\
				Surv-BART  & $3.93_{(3.93, 4.90)}$ & (0.665, 0.845) &  0.001 & (0.394, 0.517) \\
				AFT-Weibull &$\textbf{0.53}_{(0.53, 0.53)}$&(0.53,0.351)&3.088&(0.847,0.813)  \\
				AFT-log-Normal &$3.75_{(3.75, 3.75)}$&(0.717, 0.619)&7.995&(0.847, 0.321)  \\
				SR   &$0.21_{(0.21, 0.28)}$&(0.628, 0.499) &0&(1.388, 0.442)  \\
				\hdashline
				CSA (proposed)  &$0.63_{(0.59, 0.68)}$&(0.831,  0.814)& 0.132&( 1.042, 1.129)  \\
				CSA-INFO (proposed)  &$0.6_{(0.54, 0.66)}$&(0.786, 0.822)&0.13& (0.875, 0.938)  \\
			\end{tabular}
		}	
	\end{table*}

	A similar formulation for the conditional,  \emph{individualized} ${\rm HR} (t|x)$, can also be derived, where the  cumulative density $F_A(t |x) = 1- S_A(t|x)$, is estimated with a Gaussian Kernel Density Estimator (KDE) \citep{silverman1986density} on samples from the model,  $t_a  \sim p_{h, \Phi} (T_A |X=x)$. Then we have:
		\begin{definition}{Nonparametric conditional Hazard Ratio and its approximation, ${\rm \hat{HR}}(t|x)$, as}
	\begin{align}
	\begin{aligned}	
	{\rm HR} (t|x) &= \frac{\lambda_1(t|x)}{\lambda_0(t|x)} =  \frac{S_0(t|x)} {S_1(t|x)} \cdot  \frac{S_1^{\prime} (t|x)} { S_0^{\prime} (t|x)} \\
	\hat{\rm HR} (t|x) &=  \frac{\hat{S}_0^{\rm KDE}(t|x)} {\hat{S}_1^{\rm KDE}(t|x)} \cdot  \frac{m_1 (t|x)} { m_0(t|x)}\,,
	\end{aligned}
	\end{align}
\end{definition}
	where, $S^{\prime}(t|x) \triangleq d S(|t|x) /dt$ is also approximated with fitting a linear function $S(t|x) = m \cdot t + c$, and setting $S^{\prime}(t|x) \approx m$.
	Note that for some parametric formulations, ${\rm HR} (t|x)$, can be readily evaluated because $f(t_a|x)$ and $S(t_a|x)$ are available in closed form.
	
	\subsection{Factual Metrics}
	\paragraph{Concordance Index}
	C-Index (also related to receiver operating characteristic) is a widely used survival ranking metric which naturally handles censoring. It quantifies the consistency between the order of the predicted times or risk scores relative to ground truth. C-Index is evaluated on point estimates, we summarize individualized predicted samples from  CSA and CSA-INFO, \textit{i.e.}, $\hat{t}_a= {\rm median}\left(\{{t_{s}}\}^{200}_{s=1}\right)$, where $t_s$ is a sample from the trained model.
	
	\paragraph{Calibration Slope}
	Calibration quantifies distributional statistical consistency between model predictions relative to ground truth. We measure \emph{population} calibration by comparing population survival curves from model predictions against ground truth according to \citep{chapfuwa2020survival}. We desire a high calibrated model, with calibration slope of 1, while a slope $< 1$ and slope $> 1$ indicates underestimation or overestimation risk, respectively.
	
	\paragraph{Coefficient of Variation}
	The coefficient of variation (COV) $\sigma\mu^{-1}$, the ratio between standard deviation and mean, quantifies distribution dispersion. A COV $>1$ and $<1$ indicates a high or low variance distribution, in practice, we desire low variance distribution. We use Mean COV $N^{-1}\sum_{i=1}^N \sigma_i\mu_i^{-1}$, where for subject $i$ we compute $\{\mu_i, \sigma_i\}$ from samples $\{{t_{s}}\}^{200}_{s=1}$.
	
	\section{Baselines}
	
	\paragraph{Cox proportional hazard (CoxPH)} CoxPH assumes a semi-parametric linear model $\lambda(t|a) = \lambda_{\rm b}(t) \exp(a \beta)$ , thus the hazard ratio between treatment and control can be obtained without specifying the baseline hazard $\lambda_{\rm b}(t)$ as in \eqref{eq:cox-hr}.
	A simple logistic model $\hat{e}_i = \sigma(x_i; \eta)$, is used to approximate the unknown propensity score $P(A=1|X=x)$.
	Methods that adjust for selection bias (or confounding) learn $\beta$ by maximizing a propensity weighted partial likelihood \citep{schemper2009estimation, buchanan2014worth, rosenbaum1983central}
	\begin{align}
	\label{eq:coxlik}
	\mathcal{L}(\beta)  &= \prod_{i:\delta_i=1}^{} \left( \frac{\exp(a_{i} \beta)}{\sum_{j: t_{j} \ge t_{i}}  \hat{w}_j \cdot \exp(a_{j}\beta)} \right) ^{\hat{w_i}}  \,.
	\end{align}
	We consider three normalized weighting schemes for $w$ , namely, $(i)$ inverse probability weighting (IPW) \citep{horvitz1952generalization, cao2009improving}, where  ${\rm IPW}_i  = \frac{a_i}{\hat{e}_i} + \frac{1-a_i}{1-\hat{e}_i}$,  $(ii)$ overlapping weights (OW) \citep{crump2006moving, li2018balancing}, where ${\rm OW}_i = a_i \cdot (1-\hat{e}_i) + (1-a_i) \cdot \hat{e}_i$,  and  $(iii)$ the standard RCT Uniform assumption.
	Note that this modeling approach requires fitting over the entire dataset, thus has no inference capability.
	
	\paragraph{Accelerated Failure Time (AFT)}  We implement ${\rm IPM}$ regularized neural-based log-Normal and Weibull AFT baselines. Both approaches have a desirable closed form $S_{h,\Phi}(t_a|x)$, thus enabling maximum likelihood based estimation, where 
	\begin{align}
	\label{eq:aft_lik}
	-\mathcal{L}_{\rm F}^{\rm AFT}  & \triangleq   \mathbb{E}_{(y,\delta,x,a) \sim p(y,\delta, X, A)} \left[  \delta \cdot \log f_{h,\Phi}(t_a|x)  \right. \notag \\
	&+  \left. (1-\delta) \cdot \log  S_{h,\Phi}(t_a|x) \right ]\,.
	\end{align}
	
	The log-Normal mean and variance parameters are learned such that,  $\log t_a = \mu_{{h,\Phi}} (h(r, a)) + \epsilon$, where $\epsilon \sim \mathcal{N}(0, \sigma^2_{{h,\Phi}}(h(r,a)))$ and $r = \Phi (x)$. Further, we learn the Weibull scale and shape parameters, where $t_a = \lambda_{{h,\Phi}} (h(r, a)) \cdot  \left( - \log U\right)^{ \left(k_{{h,\Phi}} (h(r, a))\right)^{-1}}$ and $U \sim {\rm Uniform} (0, 1)$. We regularize \eqref{eq:aft_lik} with the  ${\rm IPM}$ loss, for maximum likelihood optimization.
	
	\paragraph{Semi-supervised regression (SR)} To demonstrate the effectiveness of our flow-based uncertainty estimation approach we contrast CSA with a deterministic accuracy objective from \citep{chapfuwa2018adversarial}, where $t_a = h(r, a)$ and:
	\begin{align}
	\label{eq:sr_lik} 
	\mathcal{L}_{\rm F}^{\rm SR} & \triangleq  \mathbb{E}_{(y,\delta,x,a) \sim p(y,\delta, X, A)} \left[ \delta \cdot \left(|y-t_a| \right)   \right.  \notag\\
	& +  \left. (1-\delta) \cdot \left(\max(0, y- t_a)\right) \right]\,,
	\end{align}
	
	where \eqref{eq:sr_lik} is regularized according to the ${\rm IPM}$ loss.
	
	\begin{figure*}[htbp]
		\begin{subfigure}[b]{.49\textwidth}
			\centering
			\includegraphics[width=.98\linewidth]{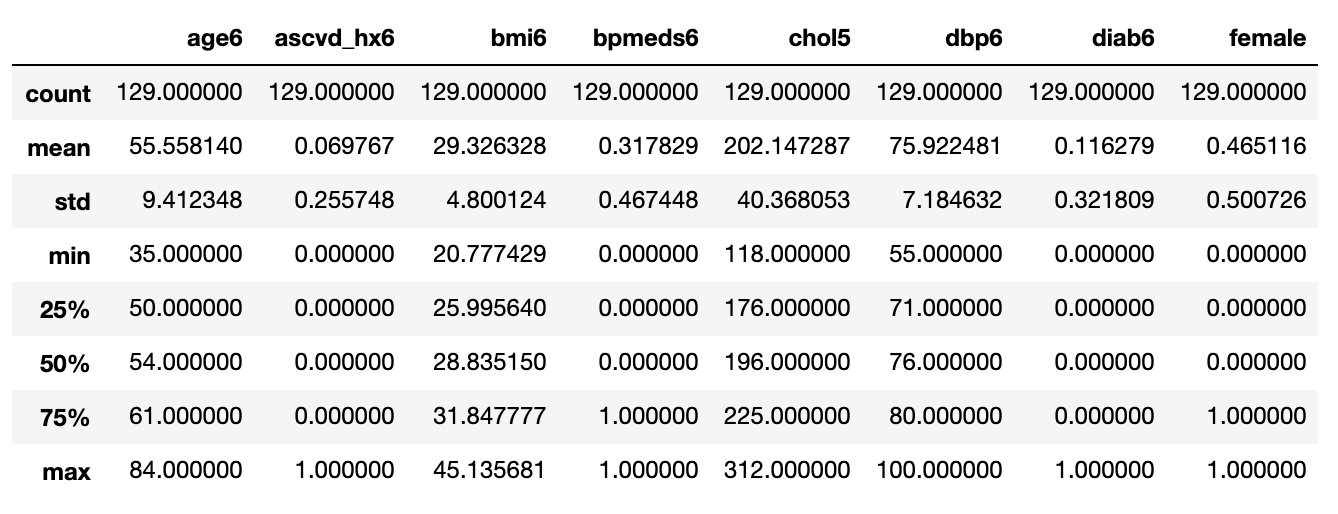}  
			\includegraphics[width=.98\linewidth]{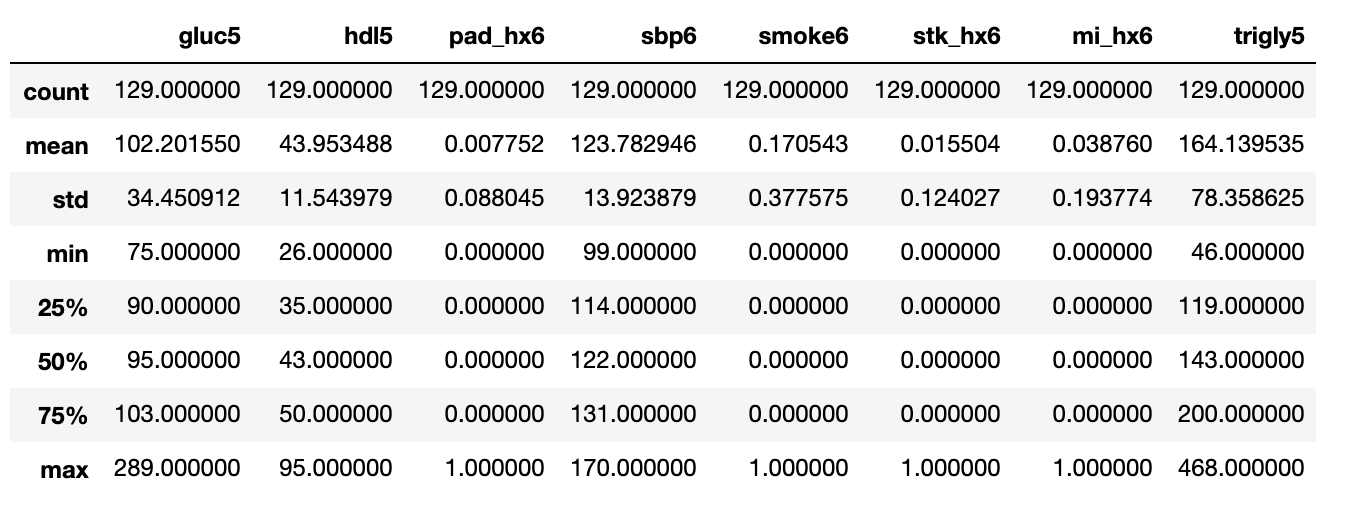}  
			\caption{ {\sc Framingham}   ${\rm HR}(t|x)<0.024$}
			\label{fig:positive}
		\end{subfigure}
		\begin{subfigure}[b]{.49\textwidth}
			\centering
			\includegraphics[width=.98\linewidth]{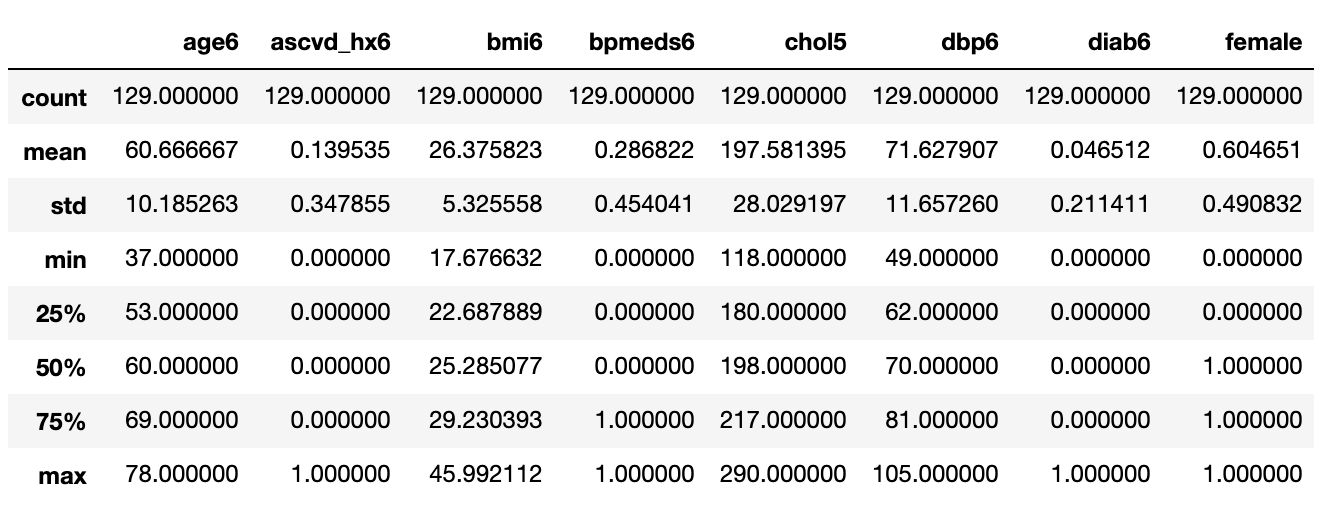}
			\includegraphics[width=.98\linewidth]{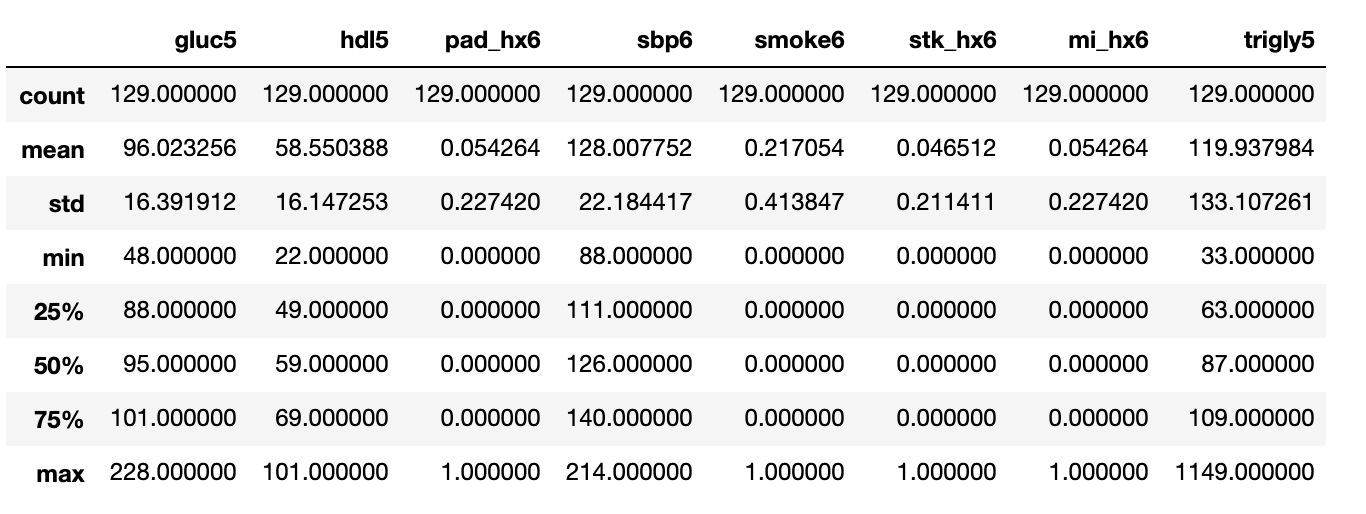}  
			\caption{ {\sc Framingham} ${\rm HR}(t|x)>1.916$ }
			\label{fig:negative}
		\end{subfigure}
		\caption{ Covariate statistics for top (a) and bottom (b) quantiles, of the median $\log {\rm HR}(t|x)$ values for the test set of {\sc Framingham}.}
		\label{fig:covariates}
	\end{figure*}

	\begin{figure}[!htpb]
	\begin{subfigure}[b]{.48\textwidth}
		\centering
		\includegraphics[width=.8\linewidth]{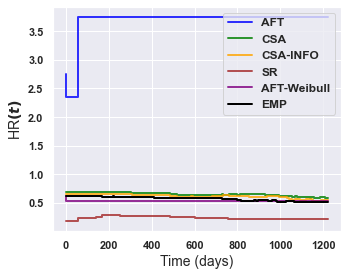}  
		\caption{ {\sc actg} ${\rm HR}(t)$}
		\label{fig:pop_hazard_actg}
	\end{subfigure}
	\begin{subfigure}[b]{.48\textwidth}
		\centering
		\includegraphics[width=.8\linewidth]{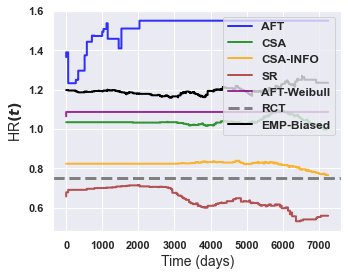}
		\caption{ {\sc Framingham} ${\rm HR}(t)$ }
		\label{fig:pop_hazard_framingham}
	\end{subfigure}
	\caption{ Inferred population ${\rm HR}(t)$ comparisons on (a) {\sc actg}  and (b) {\sc framingham}  datasets.}
	\label{fig:pop_hazards}
\end{figure}
	
	\paragraph{Survival Bayesian additive regression trees (Surv-BART)} Surv-BART \citep{sparapani2016nonparametric}  is a nonparametric tree-based  approach for estimating individualized survivals  $\hat{S} (t_a^{(j)} | X= x)$  (defined at pre-specified $J$ time-horizons)  from an ensemble of regression trees. Note, Surv-BART does not adjust for both selection bias and informative censoring. While, we fit  two separate models based on factual  treatment and control data, causal metrics are estimated with both \emph{factual} and \emph{counterfactual} predictions. 
	
	\section{Experiments}
	\subsection{Generating ATCG-Synthetic Dataset}
	The {\sc actg-Synthetic}, is a semi-synthetic dataset based on {\sc actg} covariates  \citep{hammer1996trial}.
	We simulate potential outcomes according to a Gompertz-Cox distribution \citep{bender2005generating} with selection bias from a simple logistic model for $P(A=1| X=x )$ and AFT-based censoring mechanism.  Below is our generative scheme: 
	\begin{align}
	X & = \text{ {\sc ACTG}  covariates}  \notag \\
	P(A=1|X=x) & = \frac{1}{b} \times \left(a + \sigma\left( \eta ({\rm AGE} - \mu_{\rm AGE} + {\rm CD40} - \mu_{\rm CD40}) \right) \right) \notag \\
	T_A & =  \frac{1}{\alpha_A} \log \left[1 - \frac{\alpha_A \log U}{ \lambda_A  \exp\left( x ^T  \beta_A\right)  }  \right]   \notag \\
	  U  &  \sim {\rm Uniform} (0, 1 ) \notag\\
	\log C  & \sim {\rm Normal} (\mu_c, \sigma_c^2)  \notag\\
	Y & = \min(T_A, C)  \,, \quad  \delta  =  1 \text{ if } T_A < C, \text{ else  }  \delta  = 0\notag\,,
	\end{align}
	where $\{ \beta_A, \alpha_A, \lambda_A, b, a, \eta, \mu_c, \sigma_c \}$ are hyper-parameters and $ \{\mu_{\rm AGE},  \mu_{\rm CD40}\}$ are the means for age and CD40 respectively.  This semi-synthetic  dataset will made publicly available.
	\subsection{Quantitative Results}
	
	See Table \ref{tb:actg}  for  additional quantitative comparisons on {\sc actg} dataset.

	\subsection{Qualitative Results}
	
	Figure \ref{fig:pop_hazards} demonstrates  model comparisons across of population hazard, ${\rm HR} (t)$, on {\sc actg}  and {\sc Framingham} datasets. Figure \ref{fig:covariates}, summarizes the positive and negative covariate statistics from the isolated extreme top and bottom quantiles on {\sc Framingham} datasets.
	
	\subsection {Architecture of the neural network}
	We detail the architecture of neural-based methods, namely, baselines (AFT-log-Normal, AFT-Weibull, SR) and our proposed methods (CSA and CSA-INFO). All methods are trained using one NVIDIA P100 GPU with 16GB memory.  In all experiments we set the minibatch size $M=200$, Adam  optimizer with the following hyperparameters: learning rate $3 \times 10 ^{-4}$, first moment $0.9$, second moment $0.99$, and epsilon $1 \times 10 ^{-8}$. Further, all network weights are initialed according to ${\rm Uniform}(-0.01, 0.01)$. Datasets are split into training, validation and test sets according to   70\%, 15\% and 15\%  partitions, respectively, stratified by event and treatment proportions. The validation set is used for hyperparameter search and early stopping. All hidden units in $\{h_A( \cdot), \nu_A(\cdot)\}$,  are characterized by  Leaky  Rectified Linear Unit (ReLU) activation functions, batch normalization and dropout probability of $p=0.2$ on all layers. The output layers of predicted times $\{T_A, C_A\}$ have an additional exponential transformation.
	
	\paragraph{Encoder}
	The encoding function $\Phi (\cdot)$  for mapping $r= \Phi(x)$ is shared among all the neural based methods (proposed and baselines) and specified in terms of two-layer MLPs of $100$ hidden units.
	
	\paragraph{Decoder}
	\begin{figure}[htbp]
		\begin{subfigure}[b]{.33\textwidth}
			\centering
			\includegraphics[scale=0.2]{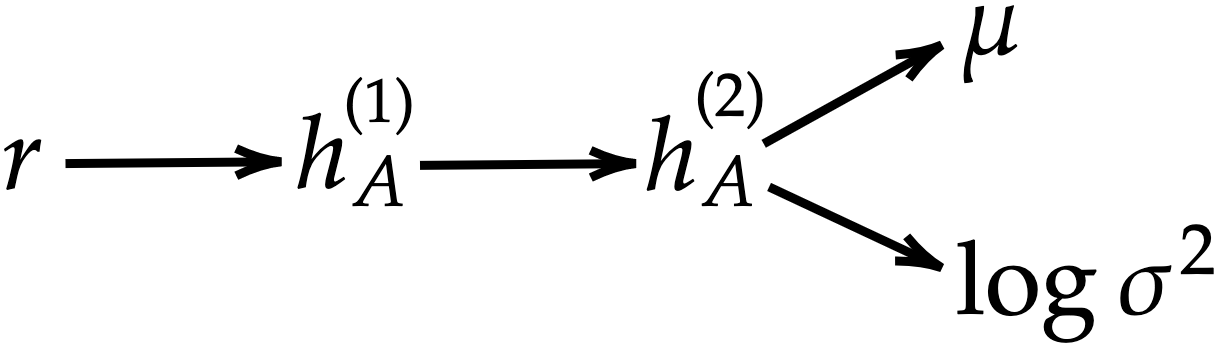}  
			\caption{ AFT-log-Normal}
			\label{fig:aft-normal}
		\end{subfigure}
		\begin{subfigure}[b]{.33\textwidth}
			\centering
			\includegraphics[scale=0.2]{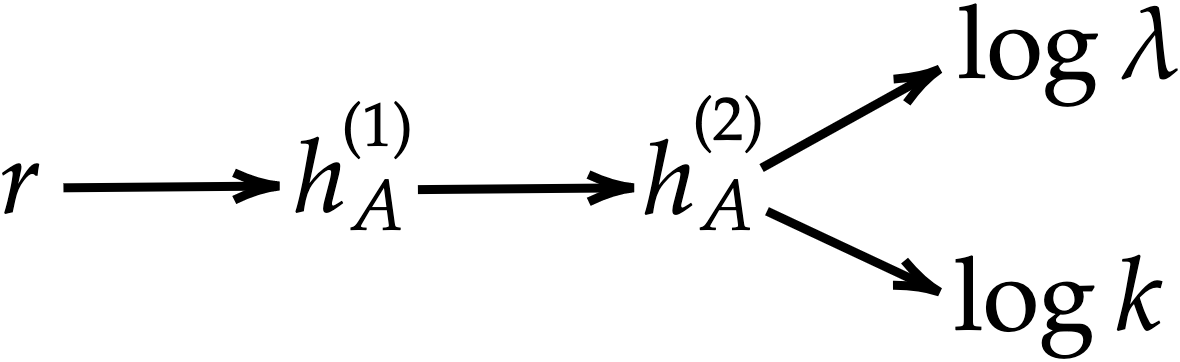}  
			\caption{AFT-Weibull}
			\label{fig:aft-weibull}
		\end{subfigure}
		\begin{subfigure}[b]{.33\textwidth}
			\centering
			\includegraphics[scale=0.2]{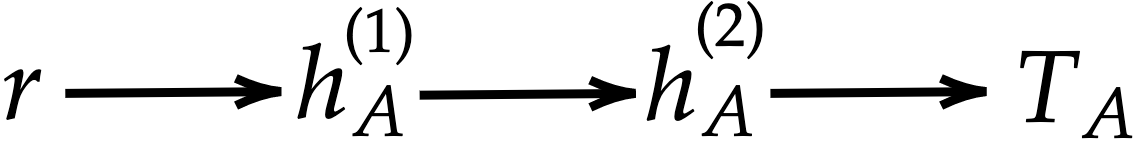}  
			\caption{ SR}
			\label{fig:sr}
		\end{subfigure}
		\caption{Decoding architecture of baselines. }
		\label{fig:baselines}
	\end{figure}
	
	\begin{figure}[htbp]
		\begin{subfigure}[b]{0.49\textwidth}
			\centering
			\includegraphics[scale=0.2]{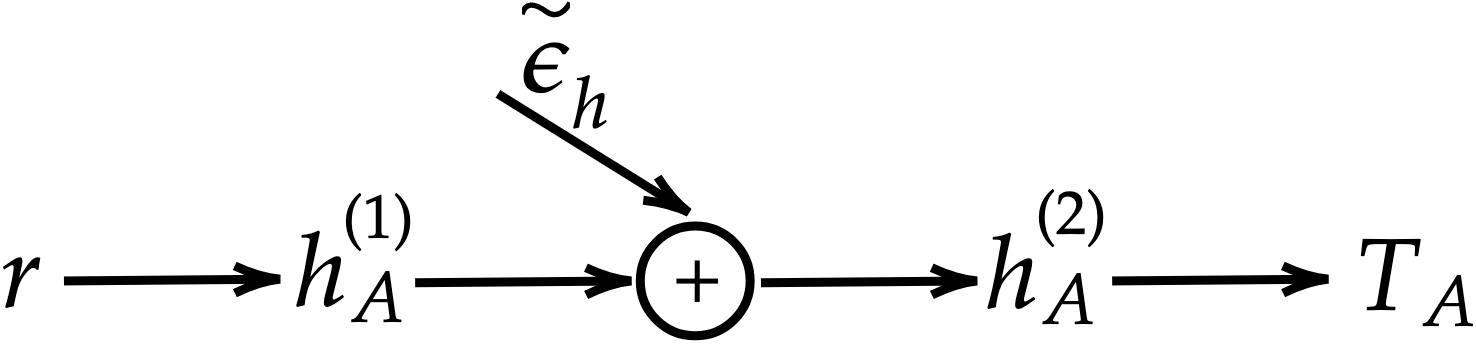}  
			\caption{ CSA}
			\label{fig:NFR}
		\end{subfigure}
		\begin{subfigure}[b]{.49\textwidth}
			\centering
			\includegraphics[scale=0.2]{figures/methods/NFR.png}  
			\includegraphics[scale=0.2]{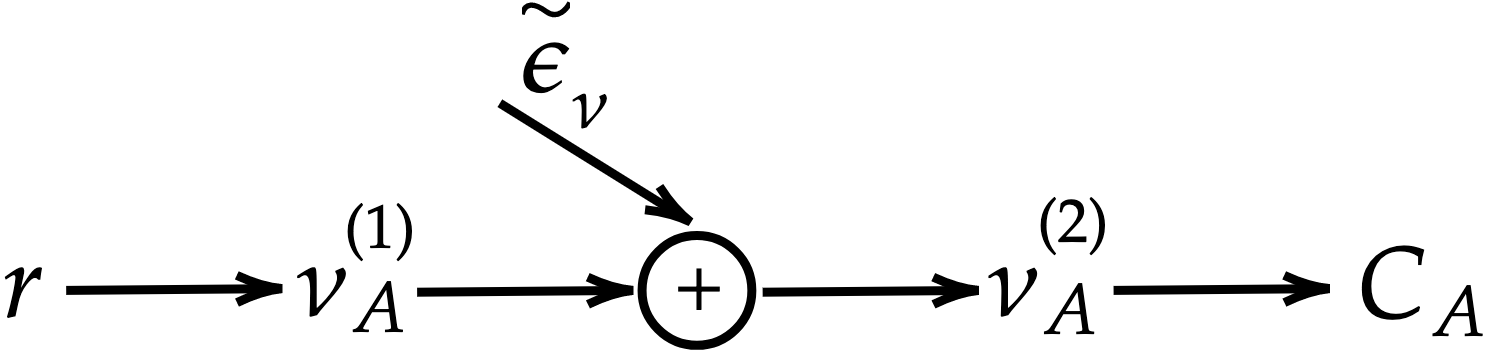}  
			\caption{CSA-INFO}
			\label{fig:NFR-INFO}
		\end{subfigure}
		\caption{Decoding architecture of  proposed methods.}
		\label{fig:proposed}
	\end{figure}

	Figure \ref{fig:baselines} shows the architectural details  of the baselines, where the decoding function $h_A(\cdot)$ is specified in terms of two-layer MLPs of $100$ hidden units.  Further, the proposed \emph{planar} flow based methods shown in Figure \ref{fig:proposed}, are comprised of two-layer MLPS for $\{h_A( \cdot), \nu_A(\cdot)\}$ of dimensions $[100, 200]$. Moreover, the hidden layers $\{h_A^{(2)}, \nu_A^{(2)}\}$, take as input the concatenated $[h_A^{(1)} , \tilde{\epsilon}_h]$ and $[\nu_A^{(1)} , \tilde{\epsilon}_\nu]$ respectively. Finally, we set the planar flow dimensions for both $\{\tilde{\epsilon}_\nu, \tilde{\epsilon}_h\}$ to $100$. 
	
